\documentclass[runningheads]{llncs}

\usepackage[mobile]{eccv}

\usepackage{eccvabbrv}

\usepackage{graphicx}
\usepackage{booktabs}

\usepackage[accsupp]{axessibility}  %

\usepackage[pagebackref,breaklinks,colorlinks,citecolor=eccvblue]{hyperref}

\usepackage{orcidlink}
\usepackage[dvipsnames]{xcolor}
\usepackage{colortbl}
\usepackage{multirow}
\usepackage{multicol}
\usepackage{svg}
\usepackage{pdfpages}
\usepackage{marvosym} %

\newcommand{\boldparagraph}[1]{\vspace{0.1cm}\noindent{\bf #1:}\looseness=-1 }

\begin{document}
\newcommand{\ba}{\mathbf{a}}\newcommand{\bA}{\mathbf{A}}
\newcommand{\bb}{\mathbf{b}}\newcommand{\bB}{\mathbf{B}}
\newcommand{\bc}{\mathbf{c}}\newcommand{\bC}{\mathbf{C}}
\newcommand{\bd}{\mathbf{d}}\newcommand{\bD}{\mathbf{D}}
\newcommand{\be}{\mathbf{e}}\newcommand{\bE}{\mathbf{E}}
\newcommand{\bff}{\mathbf{f}}\newcommand{\bF}{\mathbf{F}} %
\newcommand{\bg}{\mathbf{g}}\newcommand{\bG}{\mathbf{G}}
\newcommand{\bh}{\mathbf{h}}\newcommand{\bH}{\mathbf{H}}
\newcommand{\bi}{\mathbf{i}}\newcommand{\bI}{\mathbf{I}}
\newcommand{\bj}{\mathbf{j}}\newcommand{\bJ}{\mathbf{J}}
\newcommand{\bk}{\mathbf{k}}\newcommand{\bK}{\mathbf{K}}
\newcommand{\bl}{\mathbf{l}}\newcommand{\bL}{\mathbf{L}}
\newcommand{\bm}{\mathbf{m}}\newcommand{\bM}{\mathbf{M}}
\newcommand{\bn}{\mathbf{n}}\newcommand{\bN}{\mathbf{N}}
\newcommand{\bo}{\mathbf{o}}\newcommand{\bO}{\mathbf{O}}
\newcommand{\bp}{\mathbf{p}}\newcommand{\bP}{\mathbf{P}}
\newcommand{\bq}{\mathbf{q}}\newcommand{\bQ}{\mathbf{Q}}
\newcommand{\br}{\mathbf{r}}\newcommand{\bR}{\mathbf{R}}
\newcommand{\bs}{\mathbf{s}}\newcommand{\bS}{\mathbf{S}}
\newcommand{\bt}{\mathbf{t}}\newcommand{\bT}{\mathbf{T}}
\newcommand{\bu}{\mathbf{u}}\newcommand{\bU}{\mathbf{U}}
\newcommand{\bv}{\mathbf{v}}\newcommand{\bV}{\mathbf{V}}
\newcommand{\bw}{\mathbf{w}}\newcommand{\bW}{\mathbf{W}}
\newcommand{\bx}{\mathbf{x}}\newcommand{\bX}{\mathbf{X}}
\newcommand{\by}{\mathbf{y}}\newcommand{\bY}{\mathbf{Y}}
\newcommand{\bz}{\mathbf{z}}\newcommand{\bZ}{\mathbf{Z}}
\newcommand{\nA}{\mathbb{A}}
\newcommand{\nB}{\mathbb{B}}
\newcommand{\nC}{\mathbb{C}}
\newcommand{\nD}{\mathbb{D}}
\newcommand{\nE}{\mathbb{E}}
\newcommand{\nF}{\mathbb{F}}
\newcommand{\nG}{\mathbb{G}}
\newcommand{\nH}{\mathbb{H}}
\newcommand{\nI}{\mathbb{I}}
\newcommand{\nJ}{\mathbb{J}}
\newcommand{\nK}{\mathbb{K}}
\newcommand{\nL}{\mathbb{L}}
\newcommand{\nM}{\mathbb{M}}
\newcommand{\nN}{\mathbb{N}}
\newcommand{\nO}{\mathbb{O}}
\newcommand{\nP}{\mathbb{P}}
\newcommand{\nQ}{\mathbb{Q}}
\newcommand{\nR}{\mathbb{R}}
\newcommand{\nS}{\mathbb{S}}
\newcommand{\nT}{\mathbb{T}}
\newcommand{\nU}{\mathbb{U}}
\newcommand{\nV}{\mathbb{V}}
\newcommand{\nW}{\mathbb{W}}
\newcommand{\nX}{\mathbb{X}}
\newcommand{\nY}{\mathbb{Y}}
\newcommand{\nZ}{\mathbb{Z}}

\newcommand{\cA}{\mathcal{A}}
\newcommand{\cB}{\mathcal{B}}
\newcommand{\cC}{\mathcal{C}}
\newcommand{\cD}{\mathcal{D}}
\newcommand{\cE}{\mathcal{E}}
\newcommand{\cF}{\mathcal{F}}
\newcommand{\cG}{\mathcal{G}}
\newcommand{\cH}{\mathcal{H}}
\newcommand{\cI}{\mathcal{I}}
\newcommand{\cJ}{\mathcal{J}}
\newcommand{\cK}{\mathcal{K}}
\newcommand{\cL}{\mathcal{L}}
\newcommand{\cM}{\mathcal{M}}
\newcommand{\cN}{\mathcal{N}}
\newcommand{\cO}{\mathcal{O}}
\newcommand{\cP}{\mathcal{P}}
\newcommand{\cQ}{\mathcal{Q}}
\newcommand{\cR}{\mathcal{R}}
\newcommand{\cS}{\mathcal{S}}
\newcommand{\cT}{\mathcal{T}}
\newcommand{\cU}{\mathcal{U}}
\newcommand{\cV}{\mathcal{V}}
\newcommand{\cW}{\mathcal{W}}
\newcommand{\cX}{\mathcal{X}}
\newcommand{\cY}{\mathcal{Y}}
\newcommand{\cZ}{\mathcal{Z}}
\newcommand{\balpha}{\boldsymbol{\alpha}}\newcommand{\bAlpha}{\boldsymbol{\Alpha}}
\newcommand{\bbeta}{\boldsymbol{\beta}}\newcommand{\bBeta}{\boldsymbol{\Beta}}
\newcommand{\bgamma}{\boldsymbol{\gamma}}\newcommand{\bGamma}{\boldsymbol{\Gamma}}
\newcommand{\bdelta}{\boldsymbol{\delta}}\newcommand{\bDelta}{\boldsymbol{\Delta}}
\newcommand{\bepsilon}{\boldsymbol{\epsilon}}\newcommand{\bEpsilon}{\boldsymbol{\Epsilon}}
\newcommand{\bzeta}{\boldsymbol{\zeta}}\newcommand{\bZeta}{\boldsymbol{\Zeta}}
\newcommand{\beeta}{\boldsymbol{\eta}}\newcommand{\bEta}{\boldsymbol{\Eta}} %
\newcommand{\btheta}{\boldsymbol{\theta}}\newcommand{\bTheta}{\boldsymbol{\Theta}}
\newcommand{\biota}{\boldsymbol{\iota}}\newcommand{\bIota}{\boldsymbol{\Iota}}
\newcommand{\bkappa}{\boldsymbol{\kappa}}\newcommand{\bKappa}{\boldsymbol{\Kappa}}
\newcommand{\blambda}{\boldsymbol{\lambda}}\newcommand{\bLambda}{\boldsymbol{\Lambda}}
\newcommand{\bmu}{\boldsymbol{\mu}}\newcommand{\bMu}{\boldsymbol{\Mu}}
\newcommand{\bnu}{\boldsymbol{\nu}}\newcommand{\bNu}{\boldsymbol{\Nu}}
\newcommand{\bxi}{\boldsymbol{\xi}}\newcommand{\bXi}{\boldsymbol{\Xi}}
\newcommand{\bomikron}{\boldsymbol{\omikron}}\newcommand{\bOmikron}{\boldsymbol{\Omikron}}
\newcommand{\bpi}{\boldsymbol{\pi}}\newcommand{\bPi}{\boldsymbol{\Pi}}
\newcommand{\brho}{\boldsymbol{\rho}}\newcommand{\bRho}{\boldsymbol{\Rho}}
\newcommand{\bsigma}{\boldsymbol{\sigma}}\newcommand{\bSigma}{\boldsymbol{\Sigma}}
\newcommand{\btau}{\boldsymbol{\tau}}\newcommand{\bTau}{\boldsymbol{\Tau}}
\newcommand{\bypsilon}{\boldsymbol{\ypsilon}}\newcommand{\bYpsilon}{\boldsymbol{\Ypsilon}}
\newcommand{\bphi}{\boldsymbol{\phi}}\newcommand{\bPhi}{\boldsymbol{\Phi}}
\newcommand{\bchi}{\boldsymbol{\chi}}\newcommand{\bChi}{\boldsymbol{\Chi}}
\newcommand{\bpsi}{\boldsymbol{\psi}}\newcommand{\bPsi}{\boldsymbol{\Psi}}
\newcommand{\bomega}{\boldsymbol{\omega}}\newcommand{\bOmega}{\boldsymbol{\Omega}}
\title{REFRAME: Reflective Surface Real-Time Rendering for Mobile Devices} 

\titlerunning{REFRAME: Reflective Surface Real-Time Rendering for Mobile Devices}

\author{Chaojie Ji\inst{1} \and
Yufeng Li\inst{2}\and
Yiyi Liao\inst{1}$\textsuperscript{\Letter}$}

\authorrunning{C.Ji et al.}

\institute{Zhejiang University \and
University of Electronic Science and Technology of China\\
}

\maketitle
\renewcommand{\thefootnote}{}
\footnotetext[2]{$\textsuperscript{\Letter}$ Corresponding author.}
\renewcommand{\thefootnote}{\arabic{footnote}}

\begin{center}
    \centering
    \captionsetup{type=figure}
    \captionsetup{width=0.9\linewidth}
    \includegraphics[width=0.9\linewidth]{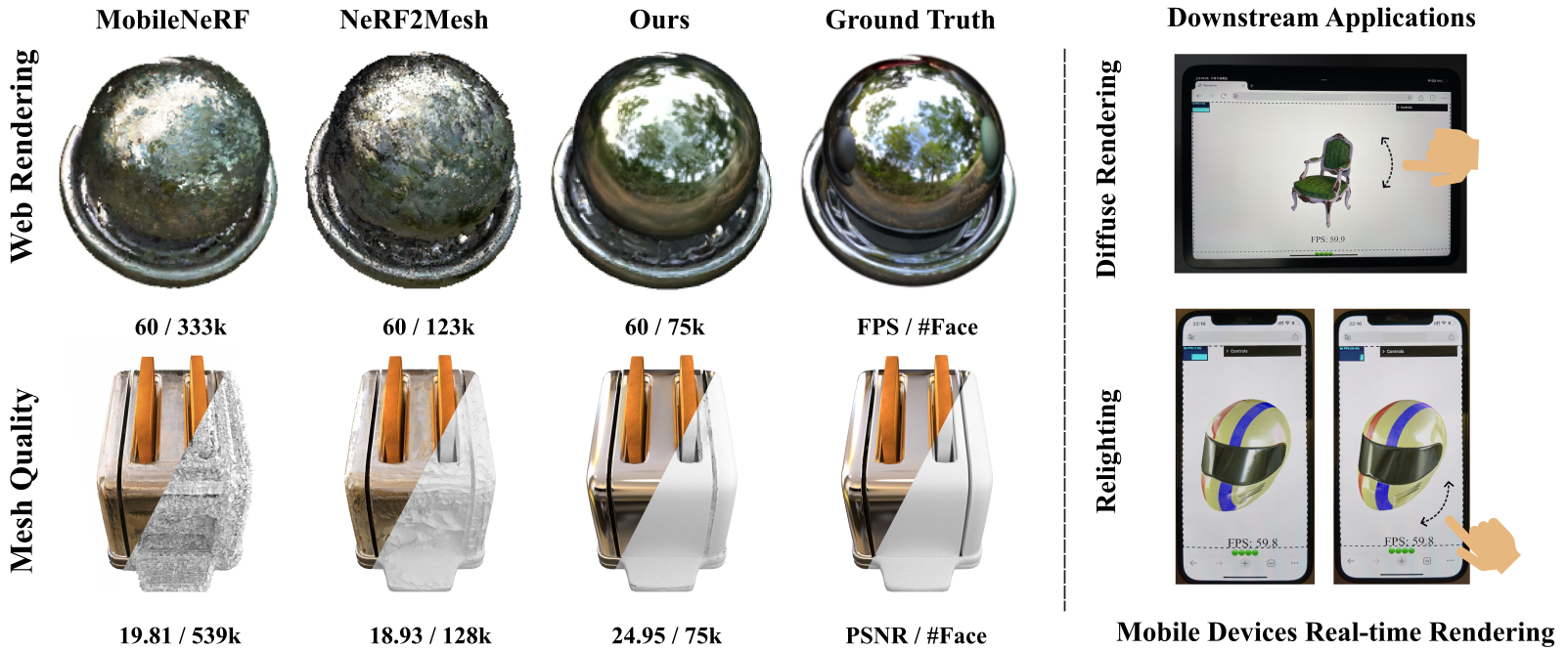}  
    \caption{\textbf{REFRAME} enables real-time rendering on consumer GPUs and mobile devices, delivering superior subjective quality with a low face number compared to the baselines~\cite{mobilenerf,nerf2mesh}. Additionally, it effectively decouples the appearance properties and environmental information of the scene, which helps its capabilities for downstream scene editing tasks.}
    \label{fig:teaser}
\end{center}%

\begin{abstract}
This work tackles the challenging task of achieving real-time novel view synthesis for reflective surfaces across various scenes.
Existing real-time rendering methods, especially those based on meshes, often have subpar performance in modeling surfaces with rich view-dependent appearances. Our key idea lies in leveraging meshes for rendering acceleration while incorporating a novel approach to parameterize view-dependent information. We decompose the color into diffuse and specular, and model the specular color in the reflected direction based on a neural environment map.
Our experiments demonstrate that our method achieves comparable reconstruction quality for highly reflective surfaces compared to state-of-the-art offline methods, while also efficiently enabling real-time rendering on edge devices such as smartphones. Our project page
is at \href{https://xdimlab.github.io/REFRAME/}{https://xdimlab.github.io/REFRAME/}.
\end{abstract}

\section{Introduction}
\label{sec:intro}

Novel view synthesis (NVS) generates realistic images from novel viewpoints using multiple input views. While Neural Radiance Fields (NeRF)~\cite{nerf} excel at high-quality NVS through volume rendering, they struggle with modeling reflective appearance and lack real-time rendering capabilities.

Several methods~\cite{nerfactor,urbaninre,tensoir} extend NeRF to decouple the intrinsic scene properties, \eg, into materials and lighting, and obtain color by the rendering equation~\cite{renderingequa}. Decoupling the environmental lighting and the physical parameters of the object often helps in modeling reflective objects. In contrast to recovering the precise physical meaning which may harm the visual quality due to the approximated rendering equation, another line of works~\cite{refnerf,refneus,denerf} avoids full decomposition but enables representing reflective objects by modeling the reflective radiance. However, none of the aforementioned methods is capable of real-time rendering, especially on mobile devices, due to the expensive query of the rendering equation, or the underlying volume rendering formulation. 

There are many existing methods focusing on accelerating the rendering of NeRF. One line of works sticks with the volume rendering pipeline where the color of a pixel is accumulated along the ray. Acceleration is achieved by improving the sampling strategy~\cite{r2l,adanerf,enerf,rtnerf}, tabulating the intermediate output~\cite{efficientnerf,fastnerf,plenoctrees,fourierplenoctrees,snerg,diver,kilonerf,instantngp}, or leveraging super-resolution neural rendering~\cite{steernerf}. However, deploying these methods on edge devices such as smartphones is often greatly hindered due to the demanding computational power they typically require. Besides, tabulation-based methods require large memory consumption and greatly increase communication costs when transmitting data between the cloud server and the client. Another line of works~\cite{nerf2mesh,nerfmeshing,mobilenerf} distills radiance fields into a mesh for real-time rendering, combined with a small MLP to model view-dependent effects. Mesh-based methods can leverage traditional graphics pipelines for acceleration, enabling them to achieve real-time rendering even on edge devices. Nonetheless, real-time rendering methods~\cite{nerf2mesh,mobilenerf,snerg,plenoctrees,vmesh} often struggle to model objects with highly reflective surfaces. Besides, these mesh-based real-time rendering methods~\cite{bakedsdf,mobilenerf,nerf2mesh} typically require a large number of vertices and faces to achieve high fidelity.

In this paper, we propose \textbf{REFRAME}, a mesh-based \textbf{REF}lective surface \textbf{R}e\textbf{A}l-time rendering method for \textbf{M}obil\textbf{E} devices (\eg, smartphones), see~\cref{fig:teaser}. We find the fact that existing NeRF distilled mesh rendering pipelines~\cite{mobilenerf,nerf2mesh} do not perform well in rendering objects with highly reflective appearances mainly for two reasons. Firstly, these methods utilize the viewing direction to model the view-dependent appearance, approved to be less effective than using the reflection direction~\cite{refnerf}.
As such, our method adopts the reflection direction-based parameterization. Nevertheless, this parameterization requires surface normal, yet estimating precise normal based on a mesh representation can be challenging. Therefore, we propose a geometry learner that learns vertex and normal offsets through two networks. This leads to good normal estimations to calculate the reflection direction while maintaining relatively low vertex and face numbers. 
Secondly and more importantly, to alleviate inference burden, real-time rendering methods often have limited capacity to model complex view-dependent information. In order to enhance the expressive power of the model without increasing the inference computational cost, we employ a four-layer MLP during training to learn the environmental feature. This information is then baked into a two-dimensional environment feature map during inference. Remarkably, the distilled environment feature map incurs a memory overhead of less than 1MB and can be edited for relighting purposes. Finally, despite our method achieving real-time rendering with an advantage in mesh faces and vertices number compared to existing works, the reconstruction quality of our method is comparable to the current state-of-the-art (including non-real-time methods) work and even outperforms them in rendering highly reflective objects. We summarize our contributions as follows:
\begin{itemize}
\item[$\bullet$] We propose a novel mesh geometry learner, allowing for robustly optimizing mesh vertex positions and normals. This leads to high rendering quality with relatively low mesh vertex and face numbers.

\item[$\bullet$] We propose to use an environment feature map to model view-dependent appearances of highly reflective objects, which enhances the capacity to reconstruct complex reflective appearances without increasing the inference burden. This further enables relighting effects.

\item[$\bullet$] Our rendering quality is on par with the current state-of-the-art methods while being able to achieve real-time rendering across various platform devices. Moreover, our method even surpasses the current non-real-time state-of-the-art approaches in rendering objects with highly reflective appearances.

\end{itemize}

\section{Related Work}
\boldparagraph{NeRF-based Scene Representation}
NeRF \cite{nerf} and its derivative works \cite{mipnerf,zipnerf,mipnerf360,humannerf,blocknerf,urbannerf,refnerf} have employed ray-marching based volume rendering methods to achieve high-quality and realistic rendering of different types of objects in various environments. 
Many follow-up works have extended NeRF in different aspects, including dynamic scene modeling~\cite{dynamicnerf,hypernerf,nerfies}, 3D-aware generation~\cite{graf,pigan,eg3d}, and semantic scene understanding~\cite{smenticnerf,panopticnerf,pnf}. In this work, we focus on addressing two limitations of NeRF, \ie, modeling of reflective objects and real-time rendering.

\begin{figure*}[!t]
  \centering
  \includegraphics[width=\linewidth]{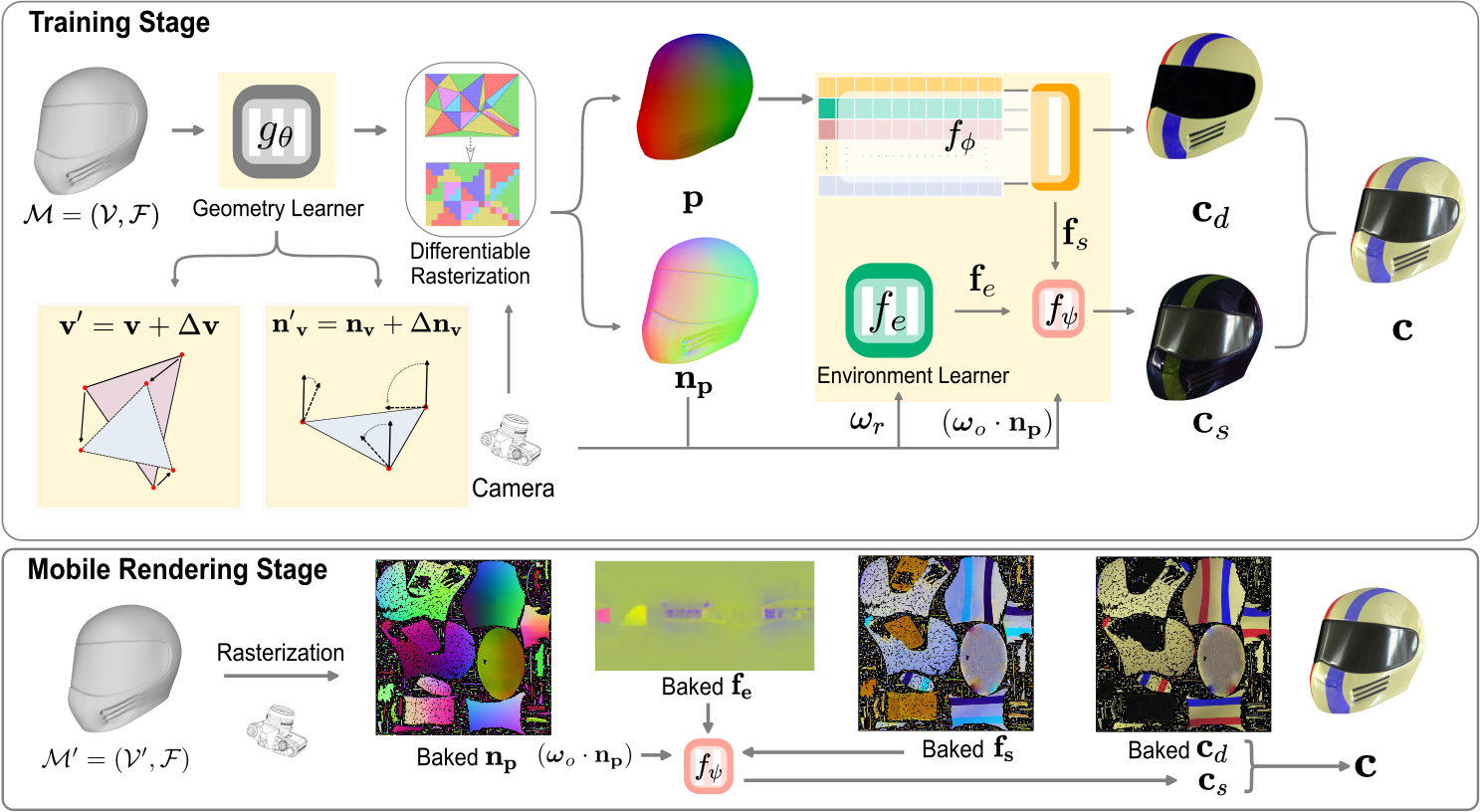}  
  \caption{\textbf{Pipeline for REFRAME.} Components with a yellow background are either baked or omitted during the mobile rendering stage. \textbf{Training Stage}: The initial mesh is updated first before performing differentiable rasterization~\cite{difras}. Next, we obtain the diffuse color $\bc_d$ and specular feature $\bff_s$ based on the position $\bp$, and the environment feature $\bff_e$ based on reflective direction $\bomega_{r}$. Then we obtain the specular color $\bc_s$ and combine it with $\bc_d$ to create the final full color $\bc$. \textbf{Mobile Rendering Stage}: We bake the intermediate output for real-time rendering. This mesh-based rendering can be implemented using WebGL and easily deployed on various platforms (\eg, desktop and mobile devices).  Here, we retrieve the $\bc_d$ and $\bff_s$ from baked texture images, and the $\bff_e$ from the environment feature map. $\bc_d$ and $\bc_s$ are processed the same as the training stage to obtain $\bc$.}
  \label{fig:pipeline}
\end{figure*}

\boldparagraph{Reflectance Decomposition} A number of works~\cite{urbaninre,iron,neilf++,nerd,neureffield,physg,nerv,nerfactor,inrmodel,tensoir,ners} investigate the task of inferring geometry and material properties based on neural fields representation, typically by formulating the image generation using the physically-based rendering. Despite achieving lighting and material control, these methods are typically inferior to state-of-the-art NeRF-based methods that directly model the radiance in terms of the rendering quality. This is due to the fact that these methods rely on simplified rendering equations of the real world.
Another line of works achieves better rendering quality by avoiding full decomposition, yet allowing for modeling glossy objects by modeling the reflected radiance~\cite{refnerf,refneus,denerf} or replacing the explicit rendering equation with a learned neural renderer~\cite{envidr,neuralpil}. However, none of these methods are capable of real-time rendering due to the underlying volume rendering formulation. While NvDiffRec~\cite{NvDiffRec} enables real-time rendering based on the mesh representation, its quality is also limited by the full decomposition. In this work, we follow the partial decomposition pipeline~\cite{refnerf} and propose to learn a neural environment map based on a NeRF-distilled mesh representation, with a focus on achieving real-time rendering.

\boldparagraph{NeRF Acceleration}
Accelerating NeRF rendering is another significant research area, with approaches reducing sampling points along the ray~\cite{adanerf,enerf,r2l}, tabulating the intermediate output~\cite{efficientnerf,fastnerf,plenoctrees,fourierplenoctrees,snerg,diver}, using thousands of small MLPs to represent the scene~\cite{kilonerf,kiloneus}, or utilizing super-resolution techniques~\cite{steernerf,4knerf}. However, these methods typically require consumer GPUs for real-time rendering, making it challenging to deploy them on edge devices with limited computational resources.

More recently, a few approaches~\cite{nerf2mesh,mobilenerf,bakedsdf} propose to distill a NeRF representation into a mesh for real-time rendering, where MobileNeRF~\cite{mobilenerf} and NeRF2Mesh~\cite{nerf2mesh} enable real-time rendering on mobile devices. However, MobileNeRF~\cite{mobilenerf} does not model the accurate geometry of the scene, leading to expensive memory costs with a large number of vertices and faces. NeRF2Mesh~\cite{nerf2mesh} is the most relevant work to ours, which distills grid-based representation into a mesh. The good geometry prior for NeRF2Mesh~\cite{nerf2mesh} leads to a relatively low number of vertices and faces. However, none of the aforementioned methods are capable of modeling highly reflective objects faithfully due to their way of view-dependent color modeling. We propose to use a neural environment map to tackle this challenge by modeling the reflected radiance. Furthermore, we propose to estimate a high-quality vertex normal, further reducing the number of faces and vertices while maintaining the quality.

\section{Method}

The pipeline of our method is illustrated in \cref{fig:pipeline}. Before training, we employ a volumetric rendering technique to obtain an initial coarse mesh similar to existing methods~\cite{nerf2mesh,mobilenerf,bakedsdf,nerfmeshing}.
During the training stage, we leverage a geometry learner to update both the vertices and the vertex normals (\cref{sec:trainingstage}). 
We further learn a shader that decomposes diffuse and specular color (\cref{sec:color_formulation}), where the specular branch is designed to be able to model highly reflective surfaces by combining view-independent specular features and reflective direction-conditioned environment features.
After training with loss functions introduced in \cref{sec:loss}, we perform UV unwrapping and bake both view-independent and view-dependent features to enable real-time rendering on various devices (\cref{sec:renderingstage}).

More formally, let $\cM=(\cV, \cF)$ denote the initial coarse mesh where $\cV=\{\bv\in\nR^3\}$ is the set of vertices and $\cF$ is the set of faces. We further compute initial vertex normals $\cN=\{\bn_\bv \in\nR^3\}$ given the initial mesh. Let $\bp \in\nR^3$ denote a rendered surface point on the mesh, $\bn_\bp \in \nR^3$ the corresponding normal, $\bomega_o \in \nS^2$ the viewing direction, and $\bomega_r  \in \nS^2$ the reflective direction. During training, our goal is to refine the mesh $\cM$, as well as map $(\bp,\bomega_r)$ to a color $\bc\in\nR^3$ to enable mesh-based real-time rendering.

\subsection{Geometry Learner}
\label{sec:trainingstage}
Before training, we follow NeRF2Mesh~\cite{nerf2mesh} to extract the initial coarse mesh through grid-based representation~\cite{instantngp}. This allows for better modeling of depth discontinuities in geometry~\cite{neus} and prevents the optimization of mesh geometry from getting stuck in local optima. Then, we propose to leverage a geometry learner to refine the geometry of the initial coarse mesh, including refining the vertex positions and normals.

\begin{figure}[t]
    \begin{subfigure}[t]{0.27\linewidth} %
        \centering
        \includegraphics[width=1.2in]{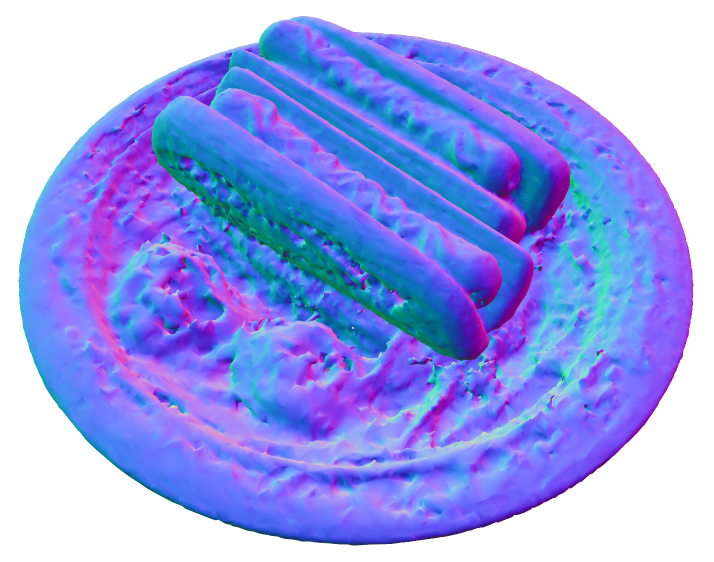} 
        \caption{Initial normal.} %
        \label{normaloptimize-a} %
    \end{subfigure}%
    \begin{subfigure}[t]{0.27\linewidth} %
        \centering
        \includegraphics[width=1.2in]{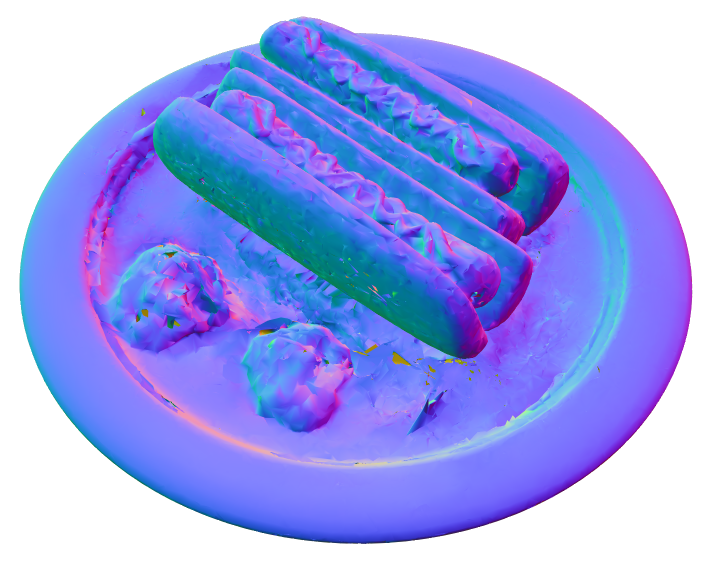} 
        \caption{Optimized normal.} %
        \label{normaloptimize-b} %
    \end{subfigure}%
    \begin{subfigure}[t]{0.23\linewidth} %
        \centering
        \includegraphics[width=1.01in]{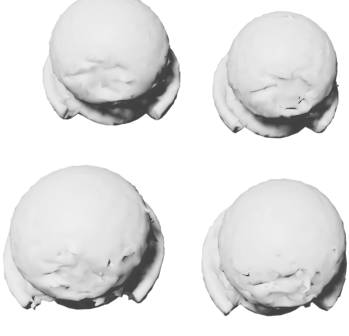} 
        \caption{Initial mesh.} %
        \label{normaloptimize-a} %
    \end{subfigure}%
    \begin{subfigure}[t]{0.23\linewidth} %
        \centering
        \includegraphics[width=1.01in]{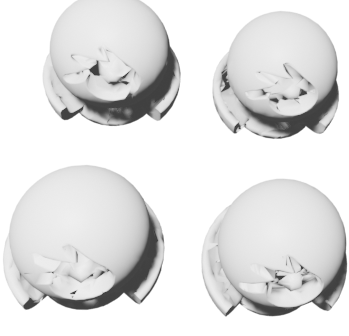} 
        \caption{Optimized mesh.} %
        \label{normaloptimize-b} %
    \end{subfigure}%
    \caption{\textbf{Mesh Optimization.} The geometry of the initial mesh is often poor. We are able to significantly improve the geometry of the mesh through geometry learner.}
  \label{fig:normaloptimize}
\end{figure}

\boldparagraph{Vertex Offset} As shown in \cref{fig:normaloptimize}, the initial mesh extracted from NeRF-based methods may be of low quality.
Existing methods \cite{nerf2mesh,nds} directly update the vertex positions through gradient backpropagation, which means the vertex update step size is determined by the learning rate and the gradient backpropagated. Thus, costly and difficult manual learning rate tuning is needed to adapt update step sizes for different objects, with a lack of robustness as shown in our experiments. To address this, our method uses a hybrid representation to provide a learned, adaptive step size for different vertices and different objects:
\begin{equation}  
\bv' = \bv + \Delta \bv, \quad  \Delta \bv = g_{{\theta}_{\bv}} (\bv)
\label{eq:geolearner_vertex}
\end{equation}  
where $\bv'$ is the updated vertex, $g_{\theta_\bv}$ is a multi-resolution hash encoding in combined with a small MLP~\cite{instantngp}. Though~\cite{mlpoff} also uses MLP for vertex offset learning, our coarse-to-fine hybrid representation allows us to leverage local information as well as global information, leading to better modeling of the geometry.

\boldparagraph{Normal Offset} Our method relies on surface normals to estimate the reflection direction $\bomega_r$ to model view-dependent appearance. Following the classical idea of smooth shading, we leverage smoothly interpolated vertex normals to approximate our surface normal. Obtaining accurate vertex normals is yet challenging, especially when the number of faces is limited.  Hence, we learn a per-vertex normal offset by taking $\bv$ and $\bn_\bv$ as input: 

\begin{equation}  
\bn'_\bv = \bn_\bv + \Delta \bn_\bv, \quad \Delta \bn_\bv = g_{\theta_\bn} (\bv,  \bn_\bv )
\label{eq:geolearner_normal}
\end{equation}  
where $\bn'_\bv$ is the learned per-vertex normal and $g_{\theta_\bn}$ is another multi-resolution hash encoding-based network.
In the experimental section, we demonstrate that learning such a per-vertex normal allows for modeling more accurate surface normals without increasing the number of vertices and faces. Note that $g_{\theta_\bv}$ and $g_{\theta_\bn}$ are only used during training, as we can directly save the updated vertices $\bv'$ and normals $\bn'_\bv$ for real-time rendering.

\subsection{Color Formulation}
\label{sec:color_formulation}

We decompose the final color $\bc$ into a diffuse color $\bc_d$ and a specular one $\bc_s$:
\begin{equation}
    \bc = \min(\max(\bc_d + \bc_s, 0), 1)
    \label{eq:color_d_s}
\end{equation}
where $\bc_d + \bc_s$ is clamped to $[0,1]$, both individual terms $\bc_d$ and $\bc_s$ from sigmoid are in [0,1]. We now elaborate on the diffuse and specular color formulation.

\boldparagraph{Diffuse Color Formulation} With the updated mesh $\cM' = (\cV', \cN', \cF)$, we perform differentiable rasterization~\cite{difras} to obtain the position $\mathbf{p}$ corresponding to each pixel. 
Subsequently, $\bp$ is mapped to a diffuse color $\bc_d \in \nR^3$ and a view-independent feature $\bff_s \in \nR^3$ which will be used later for decoding specular color:
\begin{equation}  
\bc_d, \bff_s = f_\phi(\bp)
\label{eq:hashmapping}
\end{equation}  
While the number of channels of $\bff_s$ is flexible, we observe that such a compact three-channel representation is sufficient for our purpose.

\boldparagraph{Specular Color Formulation} 
Recent non-real-time methods~\cite{refnerf,denerf,refneus} propose to model view-dependent color based on the reflective direction $\bomega_r$, instead of the viewing direction used by NeRF. This approach has been shown to effectively model reflective surfaces. Motivated by these methods, we calculate the reflective direction as follows:
\begin{equation}  
\begin{gathered}
\bomega_r = 2({\bomega_o} \cdot \bn_\bp)\bn_\bp-\bomega_o
\label{eq:refw}
\end{gathered}
\end{equation}  
where $\bn_\bp$ is smoothly interpolated surface normal at $\bp$ based on our learned vertex normal $\bn'_\bv$.
Next, the specular color can be obtained as follows:
\begin{equation}
  \bc_s = f_\psi(\bff_s,\bomega_r,({\bomega_o} \cdot \bn_\bp))
  \label{eq:cs_naive}
\end{equation}
where (${\bomega_o} \cdot \bn_\bp$) is considered as input to model the Fresnel effects~\cite{fresnel} and $f_\psi$ needs to be a tiny MLP to enable real-time rendering. 

\boldparagraph{Environment Learner}
In practice, we observe that na\"ively mapping the reflective viewing direction to specular color leads to unsatisfying performance due to the capacity limitation of the tiny MLP $f_\psi$. For a similar reason, Ref-NeRF adopts an 8-layer MLP to map $\bff_s$ and $\bomega_r$ to the view-dependent color. However, increasing the capacity of $f_\psi$ hampers the real-time rendering capability. Therefore, instead of increasing the capacity of $f_\psi$, we propose to leverage a neural environment learner of large capacity that maps $\bomega_r$ to an environment feature $\bff_e\in \nR^M$ (M=3 in our paper):

\begin{equation}  
\begin{gathered}
\bff_e = f_e(\gamma(\bomega_r)) 
\label{eq:envirlearn}
\end{gathered}
\end{equation}  
where $\gamma(\cdot)$ denotes positional encoding and $f_e$ is a 4-layar MLP.
This leads to our final implementation of the specular color:
\begin{equation}
  \bc_s = f_\psi(\bff_s, \bff_e,({\bomega_o} \cdot \bn_\bp))
  \label{eq:cs_final}
\end{equation}
Note that $\bff_e$ can be baked into a two-dimensional neural environment feature map for real-time rendering as it solely depends on $\bomega_r$.

\begin{figure}[t] %
    \begin{subfigure}[t]{0.33\linewidth} %
        \centering
        \includegraphics[width=1.25in]{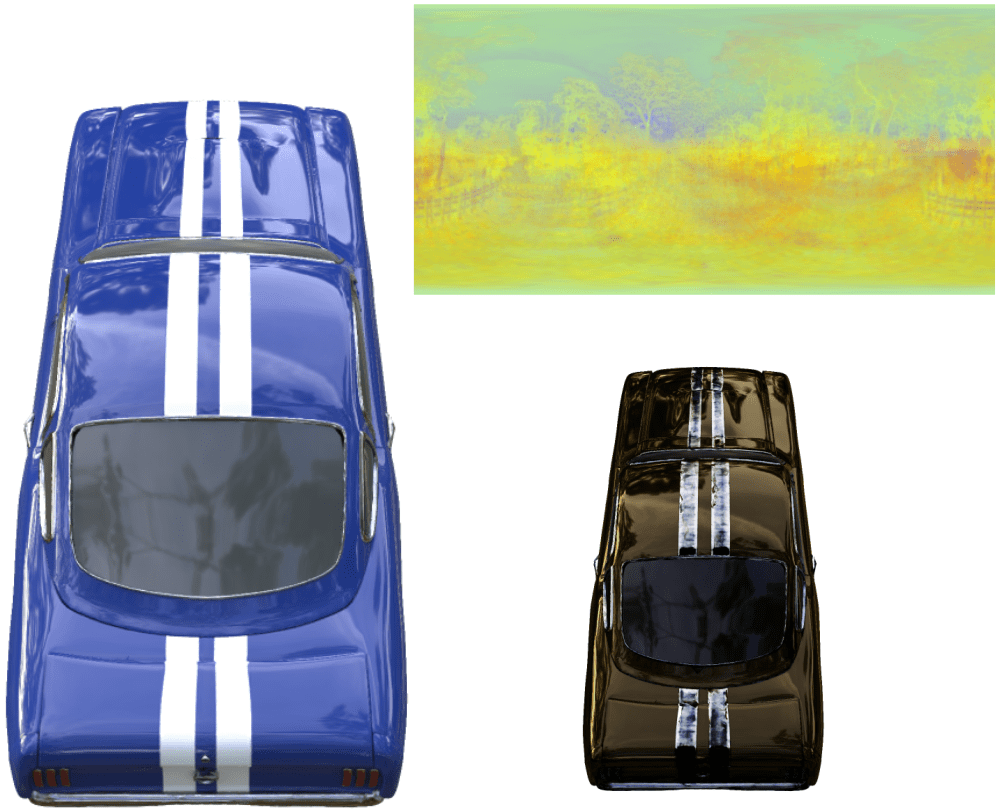} 
        \caption{Orginal image.} %
        \label{relighting-a} %
    \end{subfigure}%
    \begin{subfigure}[t]{0.33\linewidth} %
        \centering
        \includegraphics[width=1.25in]{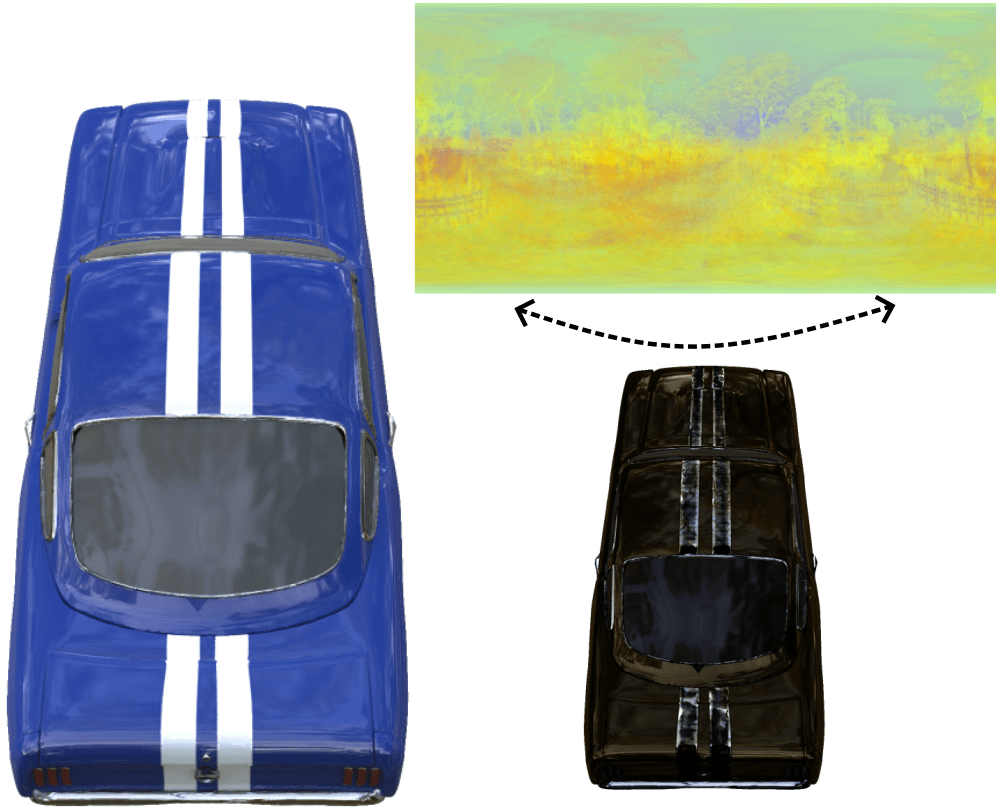} 
        \caption{Flip lighting.} %
        \label{relighting-b} %
    \end{subfigure}%
    \begin{subfigure}[t]{0.33\linewidth} %
        \centering
        \includegraphics[width=1.25in]{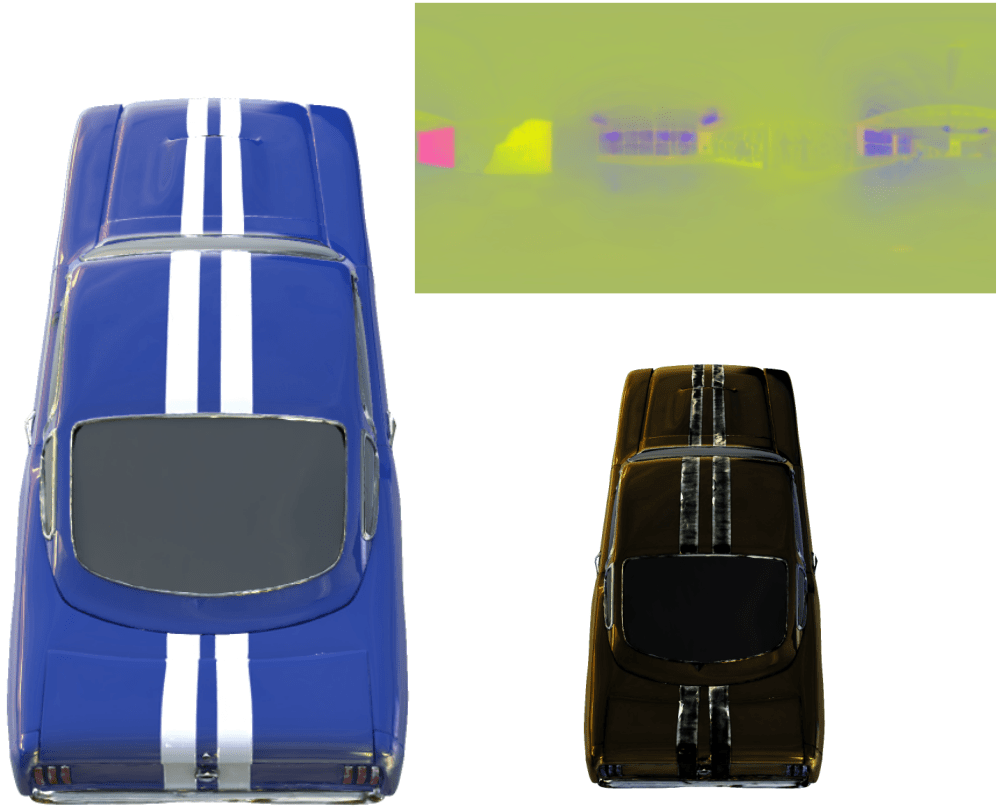} 
        \caption{Change lighting.} %
        \label{relighting-c} %
    \end{subfigure}%
    \caption{\textbf{Relighting.} We can achieve relighting of objects by editing our environment feature map. The image illustrates the editing process of flipping or replacing the environmental feature map. The left side shows the rendering result, the top right corner presents the corresponding environment feature map, and the specular color is displayed in the bottom right corner.}
  \label{fig:relighting}
\end{figure}

\boldparagraph{Scene Editing} REFRAME can decouple geometry, diffuse and specular color, so we can perform simple scene editing tasks. For example, we can edit the geometry and appearance of objects by modifying the mesh or texture image. Additionally, we can perform relighting on the objects by modifying the environment feature map of the scene, as shown in \cref{fig:relighting}.

\subsection{Loss Functions} 
\label{sec:loss}

We train our model based on image reconstruction loss and several regularization terms. 
At each iteration, we render a full image and train with the corresponding ground truth. Let $\hat{\bc}$ denote the ground truth RGB value at one pixel.
We optionally use ground truth binary mask for supervision at synthetic object-centric scenes. 
Our full loss function is shown in \cref{eq:loss}:
\begin{equation}
\begin{split}
  \mathcal{L} = & \lambda_{\bc}\mathcal{L}_{\bc} + \lambda_{\bc_d}\mathcal{L}_{\bc_d} + 
  \lambda_{SSIM}\mathcal{L}_{SSIM}  + \\
   & \lambda_{mask}\mathcal{L}_{mask} +
  \lambda_{\Delta\bn_\bv}\mathcal{L}_{\Delta\bn_\bv}+ \lambda_{max}\mathcal{L}_{max}
  \label{eq:loss}
\end{split}
\end{equation}

\boldparagraph{Image Reconstruction Loss}
Our model is first supervised by the L2 loss between the final rendered color and the ground truth.
\begin{equation}
  \mathcal{L}_{\bc} = ||\bc-\hat{\bc}||_2
  \label{eq:shadingloss}
\end{equation}

To encourage the correct decoupling of diffuse and specular colors, we add a diffuse loss to guide the diffuse color to be close to the ground truth color:
\begin{equation}
  \mathcal{L}_{\bc_d} = \|\bc_d- \hat{\bc}\|_2
  \label{eq:diffuseloss}
\end{equation}

We also introduce SSIM~\cite{ssim} loss for better perceptual quality. On object-centric datasets, we additionally utilize mask loss to ensure the rasterization mask closely aligns with the ground truth mask. Note that the mask loss is optional and is not required on unbounded datasets.

\boldparagraph{Regularizations}
We regularize the predicted vertex normal $\bn'_\bv$ to be close to the original vertex normal $\bn_\bv$ computed from the mesh. This allows for stabilizing the optimization of the vertex normals, \eg, avoid completely flipping the normal.
This is achieved by regularizing the magnitude of the normal offset:
\begin{equation}
  \mathcal{L}_{\Delta \bn_\bv} =||\Delta \bn_\bv||_1
  \label{eq:normalloss}
\end{equation}

We introduce another regularization to encourage the specular color $\bc_s$ to be reasonable when visualized alone. 
As shown in \cref{eq:color_d_s}, the final color $\bc$ will be clamped when the sum of specular color $\bc_s$ and diffuse color $\bc_d$ exceeds 1. In this case, the gradient is truncated due to the clamp operation, preventing the backpropagation from updating the color values. This may leave an undesired large specular color $\bc_s$ with no penalization.  Therefore, we encourage the sum of specular color and diffuse color to not exceed 1:
\begin{equation}
  \mathcal{L}_{max} = ||max(\bc_d+\bc_s-1, 0)||
  \label{eq:maxloss}
\end{equation}

\subsection{Real-Time Rendering}
\label{sec:renderingstage}
\boldparagraph{Environment Feature Map} During the training process, we employ a four-layer MLP with a width of 256 as our environment learner. If we query $\bff_e$ through the environment learner module during the inference process, it would significantly slow down the inference speed. Hence, during the inference phase, we bake the learned environment feature $\bff_e$ into a 2D environment feature map by converting $\bomega_r\in\nS^2$ to its polar coordinate, see supplementary for more details. This allows us to simply query the corresponding feature on the environment feature map based on $\bomega_r$, thereby greatly accelerating our inference speed, reaching above 200 FPS (frames per second) on a single NVIDIA 3090 GPU.

\boldparagraph{Texture Images} 
To further reduce the inference burden and better deploy our method across different platforms, we can also bake the $\bc_d$ and $\bff_s$ obtained from querying the $f_\phi$ into two texture images.
Specifically, we map the vertices of the mesh to UV coordinates and then bake the features into two texture images, respectively. In order to obtain the normal $\bn_\bp$ from the learned normal vector $\bn'_\bv$ during real-time rendering, we also bake the normal $\bn_\bp$ into a texture image.

\section{Experiment}

\noindent \textbf{Implementation Details.} \textbf{Training}: Prior to the training stage, we employ the classic quadric error metrics algorithm~\cite{meshsimplification} to simplify the initial mesh to 75,000 faces except for the outdoor scenes which typically have complex geometry.  We train for 250 epochs on each scene, with a training time of approximately 2 hours. We utilize the Adam optimizer~\cite{adam} and employ the cosine annealing learning rate adjustment strategy~\cite{cos} during the training stage. \textbf{Rendering}: The resolution of our baked texture image is $4096 \times 4096$, and the resolution of the baked environment feature map is $720 \times 360$. As referenced in~\cref{sec:renderingstage}, once we have baked the environment learner into the environment feature map, our approach is already capable of achieving above 200 FPS on the GPU (referred to as Ours). This implementation ensures a fair comparison with the baselines since their results quoted are evaluated using GPU implementation that can not directly deploy on mobile devices. Given that we use texture images for rendering when implemented on mobile devices, we also employ texture images on the GPU for quality evaluation (referred to as Ours (Mobile)). Following NeRF2Mesh~\cite{nerf2mesh} and MobileNeRF~\cite{mobilenerf}, we employ anti-aliasing to enhance our rendering quality. For details on the implementation of anti-aliasing, please refer to the supplementary material.
All experiments are performed on a single NVIDIA 3090 GPU.

\noindent \textbf{Dataset.} We validate the effectiveness and robustness of our method using three different datasets with varied scenes. 1) \textbf{NeRF~\cite{nerf} Synthetic Dataset}: This dataset consists of eight synthetic scenes. 
2) \textbf{Shiny Blender Dataset}: This dataset is introduced by Ref-NeRF~\cite{refnerf} and includes six glossy objects, serving as the primary dataset for assessing the reflective surface modeling capabilities of the methods.
3) \textbf{Real Captured Dataset}: We select three outdoor scenes with rich reflective appearances same as Ref-NeRF~\cite{refnerf} to validate the effectiveness of our method in real-world outdoor capture scenarios. Due to the poor quality of the initial mesh extracted by Nerf2Mesh~\cite{nerf2mesh} when modeling glossy objects, we utilize the meshes extracted by Neuralangelo~\cite{neuralangelo} and Ref-NeuS~\cite{refneus} as our initialization in certain scenes, see supplementary for more details.

\noindent \textbf{Baseline.} We compare with various types of baselines, including Ref-NeRF~\cite{refnerf}, 3DGS~\cite{3DGS}, NvDiffRec~\cite{NvDiffRec}, MobileNeRF~\cite{mobilenerf}, and NeRF2Mesh~\cite{nerf2mesh} in our study.  Ref-NeRF~\cite{refnerf} serves as the state-of-the-art volume rendering method for handling the strong reflective appearance. 3DGS~\cite{3DGS} represents a powerful new representation capable of real-time rendering on GPU with great expressive power. NvDiffRec~\cite{NvDiffRec} represents physically based rendering methods. MobileNeRF~\cite{mobilenerf}, and NeRF2Mesh~\cite{nerf2mesh} are representative works in the field of real-time rendering for mobile devices. 
We import the results of the baselines from their original paper and replicate the results that missing in their original paper.

\subsection{Reconstruction Quality}
\label{sec:reconstructionquality}

\begin{table*}[t]
\small
  \centering
 \resizebox{!}{2.0cm}{
   \begin{tabular}{c|ccc|ccc|ccc}
    \toprule
          & \multicolumn{3}{c|}{\textcolor[rgb]{ .2,  .2,  .2}{NeRF Synthetic Dataset}} & \multicolumn{3}{c|}{\textcolor[rgb]{ .2,  .2,  .2}{Shiny Blender Dataset}} & \multicolumn{3}{c}{\textcolor[rgb]{ .2,  .2,  .2}{Real Captured Dataset}} \\
          & PSNR$\uparrow$  & SSIM$\uparrow$  & LPIPS$\downarrow$ & PSNR$\uparrow$  & SSIM$\uparrow$  & LPIPS$\downarrow$ & PSNR$\uparrow$  & SSIM$\uparrow$  & LPIPS$\downarrow$ \\
    \hline
    NeRF~\cite{nerf}  & 31.01 & 0.947 & 0.081 & -     & -     & -     & -     & -     & - \\
    Ref-NeRF~\cite{refnerf} & \cellcolor[rgb]{ 1,  .8,  .8}33.99 & \cellcolor[rgb]{ .973,  .796,  .678}0.966 & \cellcolor[rgb]{ .973,  .796,  .678}0.038 & \cellcolor[rgb]{ .973,  .796,  .678}35.96 & \cellcolor[rgb]{ 1,  .949,  .8}0.967 & \cellcolor[rgb]{ 1,  .949,  .8}0.058 & \cellcolor[rgb]{ 1,  .8,  .8}24.45 & \cellcolor[rgb]{ 1,  .8,  .8}0.665 & \cellcolor[rgb]{ 1,  .8,  .8}0.142 \\
    \hline
    3DGS~\cite{3DGS}  & \cellcolor[rgb]{ .973,  .796,  .678}33.30 & \cellcolor[rgb]{ 1,  .8,  .8}0.969 & \cellcolor[rgb]{ 1,  .8,  .8}0.030 & 30.37 & 0.947 & 0.083 & \cellcolor[rgb]{ .973,  .796,  .678}24.06 & \cellcolor[rgb]{ .973,  .796,  .678}0.661 & \cellcolor[rgb]{ .973,  .796,  .678}0.259 \\
    NvDiffRec~\cite{NvDiffRec} & 29.05 & 0.939 & 0.081 & 29.05 & 0.938 & 0.111 & -     & -     & - \\
    MobileNeRF~\cite{mobilenerf} & 30.90 & 0.947 & 0.062 &  26.62  &  0.883  &  0.163     & \cellcolor[rgb]{ 1,  .949,  .8}22.96 & 0.493 & 0.430 \\
    NeRF2Mesh~\cite{nerf2mesh} & 29.67 & 0.940 & 0.072 & 26.96 & 0.896 & 0.170 & 22.71 & 0.523 & 0.419 \\
    Ours & \cellcolor[rgb]{ 1,  .949,  .8}31.04 & \cellcolor[rgb]{ 1,  .949,  .8}0.961 & \cellcolor[rgb]{ 1,  .949,  .8}0.040 & \cellcolor[rgb]{ 1,  .8,  .8}36.02 & \cellcolor[rgb]{ 1,  .8,  .8}0.983 & \cellcolor[rgb]{ 1,  .8,  .8}0.040 & 22.50 & \cellcolor[rgb]{ 1,  .949,  .8}0.598 & \cellcolor[rgb]{ 1,  .949,  .8}0.325 \\
    \hline
    Ours (Mobile) & 30.84 & 0.959 & 0.042 & \cellcolor[rgb]{ 1,  .949,  .8}35.83 & \cellcolor[rgb]{ .973,  .796,  .678}0.981 & \cellcolor[rgb]{ .973,  .796,  .678}0.041 & 22.11 & 0.524 & 0.423 \\
    \bottomrule
    \end{tabular}%
    }
    \vspace{3pt}
    
     \caption{\textbf{Rendering Quality.} Baseline comparisons of the rendering quality on three different datasets. Red represents the optimal, orange represents the second best, and yellow represents the third.}
  \label{tab:renderingquality}%
\end{table*}%

\begin{figure}[thb]
  \centering
  \begin{subfigure}{0.38\linewidth}
    \includegraphics[width=1\linewidth]{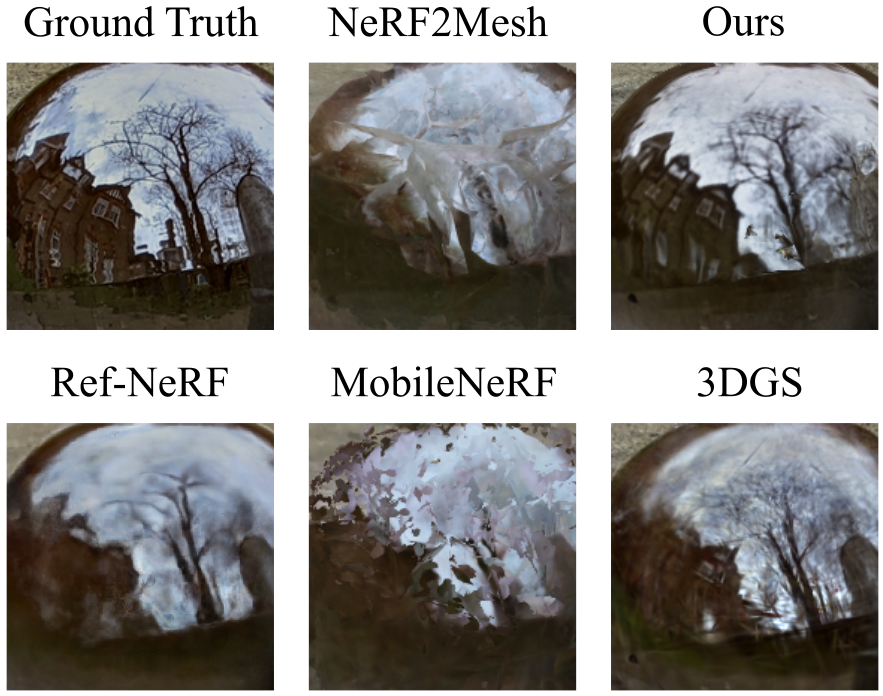}
    \caption{\textbf{Rendering Quality} on Real Captured Dataset. }
    \label{fig:gardensphere}
  \end{subfigure}
  \hfill
  \begin{subfigure}{0.60\linewidth}
    \includegraphics[width=1\linewidth]{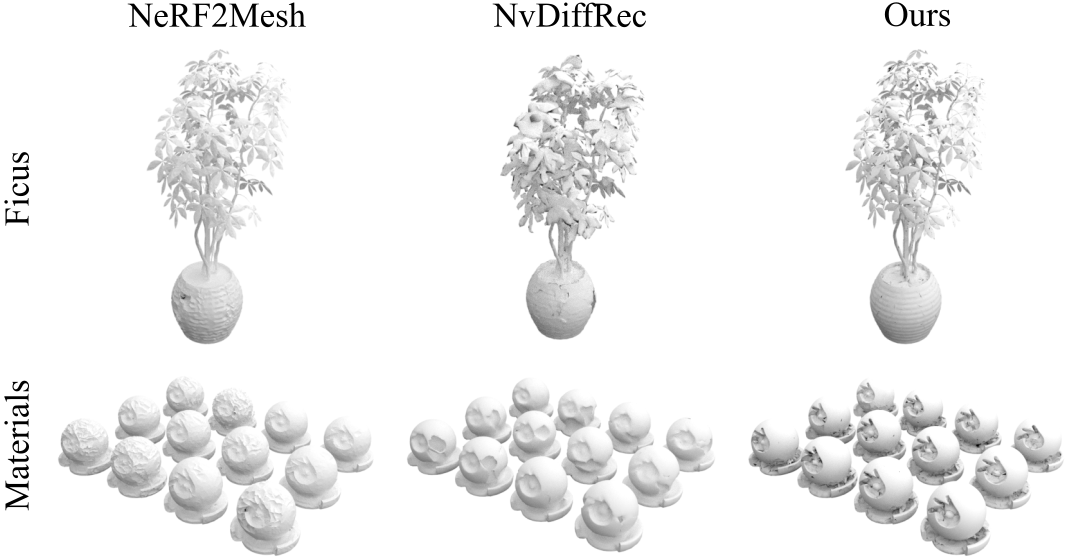}
    \caption{\textbf{Surface Quality} on NeRF Synthetic Dataset. }
    \label{fig:surfacequality}
  \end{subfigure}
  \vspace{-5pt}
  \caption{\textbf{Qualitative Comparison.} Compared to the baselines, our method reconstructs reflective regions of higher fidelity and yields more accurate surface geometry.}
  \label{fig:recon_quality}
\end{figure}

\noindent \textbf{Rendering Quality.} We compare the rendering quality of our method and baseline methods on three datasets, see~\cref{tab:renderingquality}. It is worth noting that Ref-NeRF~\cite{refnerf} is the state-of-the-art method for modeling glossy objects but cannot achieve real-time rendering, even on high-end GPU. We introduce Ref-NeRF~\cite{refnerf} as a benchmark for rendering quality and demonstrate that our method achieves comparable or even superior rendering quality to non-real-time state-of-the-art methods. While 3DGS~\cite{3DGS} can render in real-time on GPU and has higher PSNR on both NeRF Synthetic and Real Captured dataset than ours, it's important to note that 3DGS struggles to model reflective surfaces effectively (\cref{fig:gardensphere}). Moreover, they have a larger memory overhead than us. \Eg, for gardenspheres, the memory cost for 3DGS is 1.4GB, whereas our method only requires 84.3MB.

\begin{figure*}[t]
  \centering
   \includegraphics[width=1\linewidth]{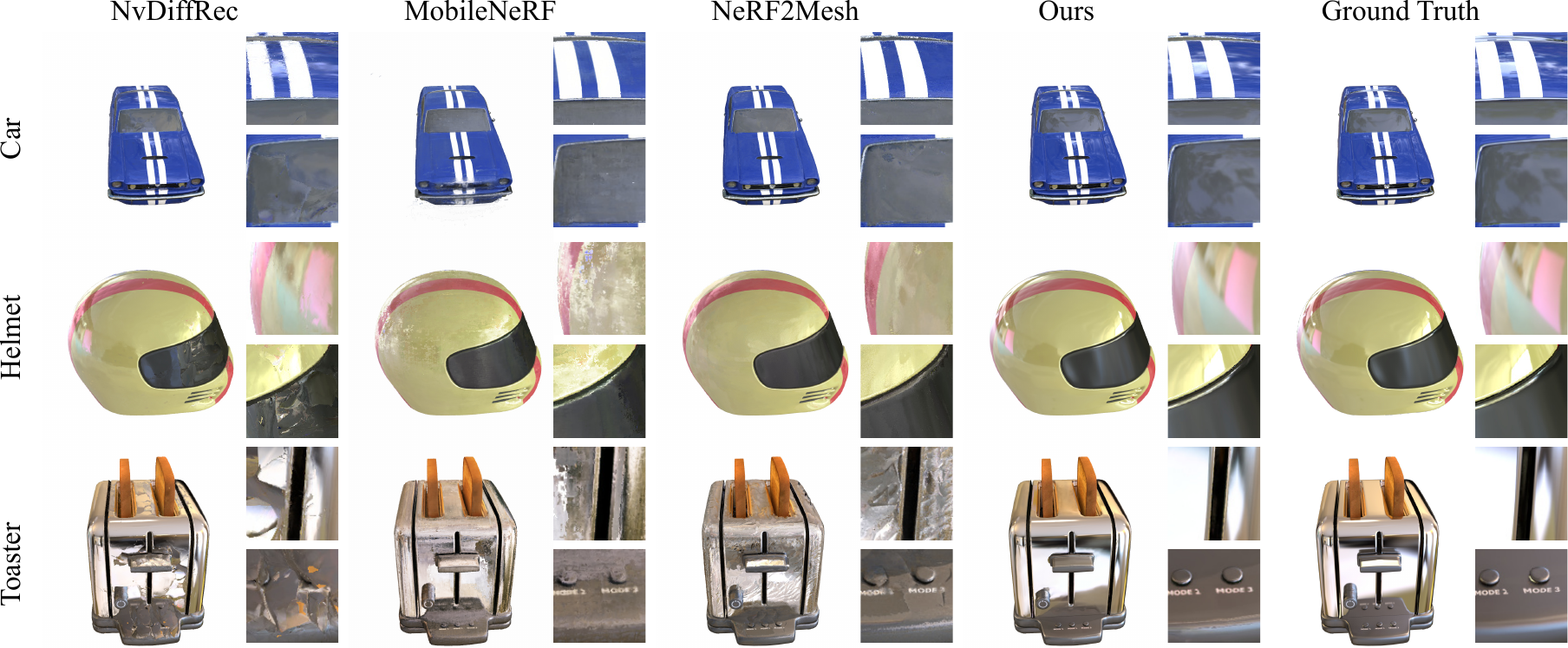}
   \vspace{-10pt}
   \caption{\textbf{Rendering Quality on Shiny Blender Dataset.} Our method achieves optimal rendering quality in most scenes and provides better modeling of reflective appearance compared to the comparison methods.}
   \label{fig:renderingquality}
\end{figure*}

In comparison to mesh-based real-time rendering methods~\cite{nerf2mesh,mobilenerf,NvDiffRec}, we achieve optimal rendering quality in most object-centric datasets, as shown in~\cref{fig:renderingquality}. Despite having mask supervision on object-centric datasets as well, NvDiffRec~\cite{NvDiffRec} and NeRF2Mesh~\cite{nerf2mesh} struggle to achieve high rendering quality. This could be due to the fact that NvDiffRec relies on the simplified rendering equation, whereas NeRF2Mesh's simple color formulation struggles to model highly reflective objects. Note that NeRF2Mesh also requires an initial mesh in its second stage, and we provide NeRF2Mesh with the same initial mesh as ours for all experiments for a fair comparison.  

In unbounded outdoor scenes without masks, our method still achieves high-quality rendering, as demonstrated in~\cref{fig:gardensphere}. Although our PSNR may be slightly lower compared to the baselines, we outperform them in rendering foreground glossy objects. However, this advantage is not accurately reflected in the PSNR metric since the foreground reflective objects occupy only a small region in the image.
Further, our method mainly exhibits artifacts in the background, this is due to the fact that the initial mesh can only faithfully reconstruct the foreground but not the background, see supplementary for more details.

\noindent \textbf{Surface Quality.} Furthermore, we observe that our method not only improves the rendering quality when modeling glossy objects but also generates more accurate meshes, as shown in~\cref{fig:surfacequality}. It can be seen that the meshes obtained from~\cite{nerf2mesh,NvDiffRec} often have poor quality in the reflective region while our method is capable of accurately reconstructing the geometry.

\subsection{Rendering Efficiency}
\label{sec:renderingefficiency}
\begin{table}[t]
  \centering
    \small
    
    \begin{tabular}{c|ccccc}
    \toprule
          & \multicolumn{1}{l}{MacBook Pro} & \multicolumn{1}{l}{iPhone12} & \multicolumn{1}{l}{iPad Air3} & \multicolumn{1}{l}{Legion Y-7000} & \multicolumn{1}{l}{HUAWEI nova7} \\
    \hline
    MobileNeRF~\cite{mobilenerf} & 120.00   & 60.00    & 60.00    & 55.31 & 39.19 \\
    NeRF2Mesh~\cite{nerf2mesh} & 120.00   & 60.00    & 60.00    & 48.69 & 32.56 \\
    Ours (Mobile)  & 120.00   & 60.00    & 60.00    & 48.50  & 31.71 \\
    \bottomrule
    \end{tabular}%
    \vspace{3pt}
    
    \caption{\textbf{Computational Efficiency Comparison.} We test the FPS of our method and the baseline methods~\cite{nerf2mesh,mobilenerf} on the NeRF Synthetic dataset across multiple devices.}
  \label{tab:fps}%
\end{table}%

\noindent \textbf{Computational efficiency.} During the mobile rendering stage, the bottleneck of our rendering speed lies in querying the tiny MLP $f_\psi$ to obtain the specular color. However, due to our MLP being only two layers with 64 widths, our rendering speed remains relatively fast. Therefore, our method achieves real-time rendering on both consumer GPUs and mobile devices. We compare the rendering speed with the baselines~\cite{nerf2mesh,mobilenerf} that can render in real-time on mobile devices across multiple platform devices, as seen in~\cref{tab:fps}.

\noindent \textbf{Memory efficiency.} Our cache mainly consists of texture images, the environment feature map, and the mesh. Due to our high-quality estimation of vertex normals, our mesh can achieve high-quality rendering with a low number of vertices and faces, leading to a small cache size in most scenes, as shown in~\cref{tab:memory}. 
\begin{table}[htbp]
  \small
      \setlength{\tabcolsep}{2mm}
  \centering
    \begin{tabular}{c|ccc|ccc}
    \toprule
          & \multicolumn{3}{c|}{NeRF Synthetic Dataset} & \multicolumn{3}{c}{Shiny Blender Dataset} \\
          & \#V ($10^3$)   & \#F ($10^3$)   & Cache & \#V ($10^3$)   & \#F ($10^3$)   & Cache \\
    \hline
    MobileNeRF~\cite{mobilenerf} & 494   & 224   & 125.75 & 1028  & 343   & 152.37 \\
    NeRF2Mesh~\cite{nerf2mesh} & 200   & 192   & 73.53 & 45    & 91    & 22.15 \\
    Ours (Mobile) & 37    & 75    & 47.55 & 38    & 75    & 52.86 \\
    \bottomrule
    \end{tabular}%
    \vspace{3pt}
    
    \caption{\textbf{Memory Efficiency.} While exhibiting higher quality, our method maintains a relatively low memory consumption compared to baselines. Cache reported in MB.}
  \label{tab:memory}%
\end{table}%

\begin{table}[h]
\begin{minipage}{.48\linewidth}

\centering
        \resizebox{!}{0.9cm}{
\begin{tabular}{c|cc|c}
    \toprule
          & \multicolumn{2}{c|}{w/o $g_{{\theta}_{\bv}}$} & Ours \\
          & L=0.001 & L=0.0001 & L=0.001 \\
    \hline
    Toaster & 17.01 & 17.01 & 24.95 \\
    Materials & 23.01 & 29.06 & 29.58 \\
    \bottomrule
    \end{tabular}%
    }
    \vspace{3pt}
    
    \caption{\textbf{Ablation of Vertex Offset Learning}. L represents the learning rate. Reported in PSNR.}
    \label{tab:ablveofftab}%

\end{minipage}%
\hfill
\begin{minipage}{.48\linewidth}

\centering
        \resizebox{!}{1.1cm}{
\begin{tabular}{cc|cc}
    \toprule
          &       & \multicolumn{1}{l}{w/o $\Delta \bn_\bv$} & \multicolumn{1}{l}{Ours} \\
    \hline
    \multirow{2}[1]{*}{Toaster} & \#F=10k & 24.24 & 24.62 \\
          & \#F=75k & 24.55 & 24.95 \\
    \multirow{2}[1]{*}{Materials} & \#F=10k & 27.69 & 28.23 \\
          & \#F=75k & 29.46 & 29.58 \\
    
    \bottomrule
    \end{tabular}%

    }
    \vspace{3pt}
    
        \caption{\textbf{Ablation of Normal Offset Learning}. Reported in PSNR.}
        \label{tab:ablnorofftab}%

\end{minipage}
\end{table}

\subsection{Ablation Study}
\label{sec:ablationstudy}

\noindent \textbf{Geometry Learner.} ~\cite{nds} and ~\cite{nerf2mesh} utilize direct gradient backpropagation to update the vertex positions of the mesh. We employ this update strategy in our method and verify that this update method lacks robustness. Directly using the backpropagated gradients to update the vertex positions can lead to a devastating blow to the mesh, especially when using a large learning rate. This significantly compromises the rendering quality, as shown in~\cref{tab:ablveofftab}. Additionally, different datasets require different learning rates, as indicated in~\cref{tab:ablveofftab}. While a learning rate of 0.0001 works fine for the materials dataset, it performs poorly on the toaster dataset. In contrast, our vertex update strategy not only exhibits robustness across different datasets but also provides higher rendering quality.

In order to verify the necessity of estimating an offset for the vertices normal, we conduct experiments with the w/o normal offset setting. In this setting, we do not estimate a normal offset for each vertex, but we still compute an updated normal based on the updated vertex positions. As shown in~\cref{tab:ablnorofftab}, introducing per-vertex normal offset provides a gain in reconstruction quality. This gain is particularly significant when the number of faces is low. Notably, in the case of the toaster dataset, a mesh with 10k faces with normal offset even achieves rendering quality surpassing that of a mesh with 75k faces without normal offset.

\begin{figure}[t]
    \begin{minipage}{0.63\textwidth}
        \centering
    \begin{subfigure}[t]{0.33\linewidth} %
        \centering
        \includegraphics[width=0.75in]{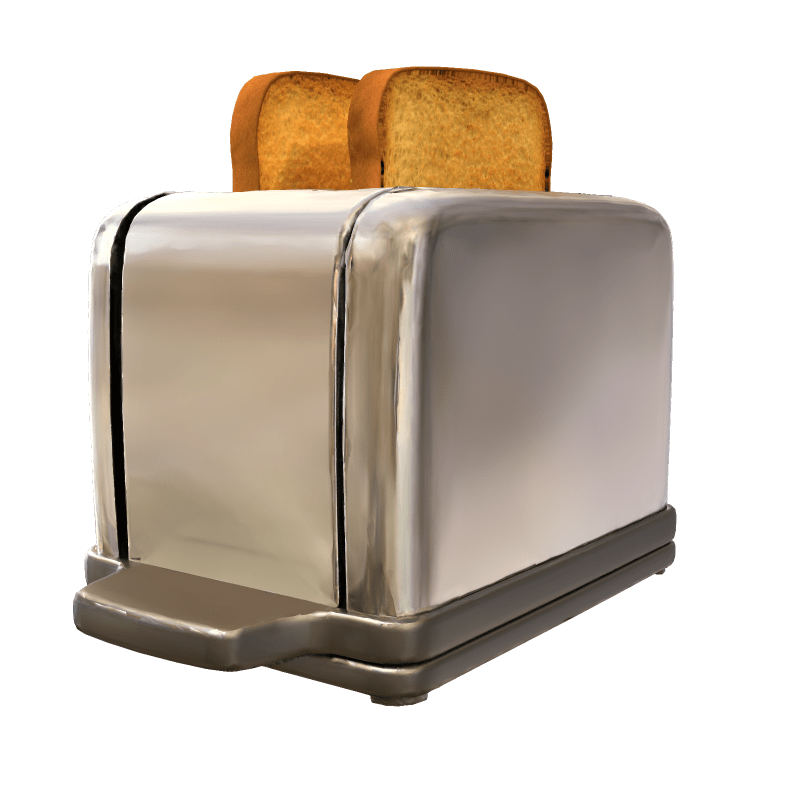} 
        \caption{w/o $f_e$} %
    \end{subfigure}%
    \hfill
    \begin{subfigure}[t]{0.33\linewidth} %
        \centering
        \includegraphics[width=0.75in]{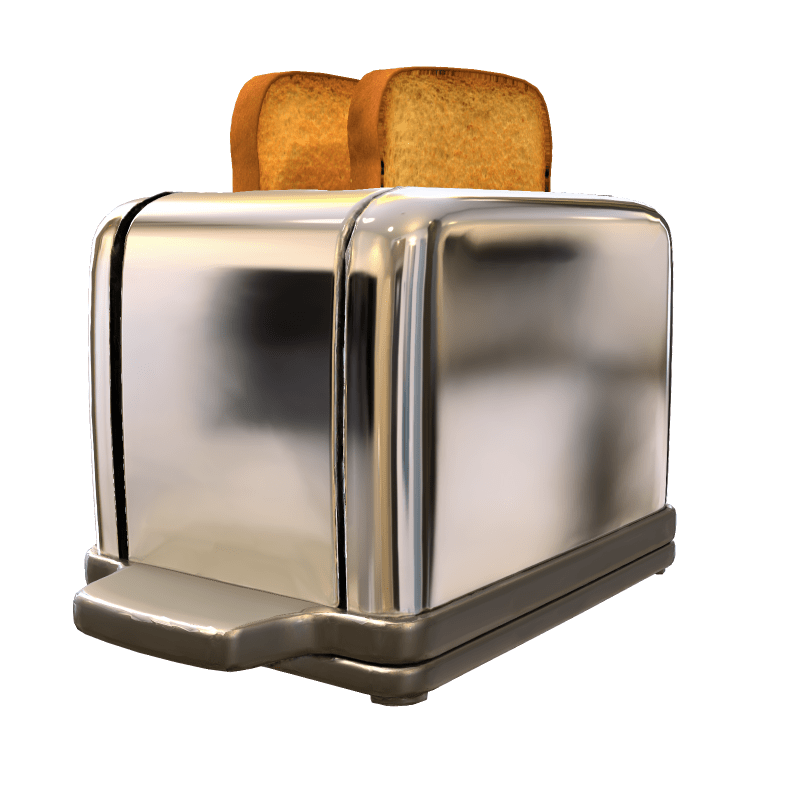} 
        \caption{Ours} %
    \end{subfigure}%
    \begin{subfigure}[t]{0.33\linewidth} %
        \centering
        \includegraphics[width=0.75in]{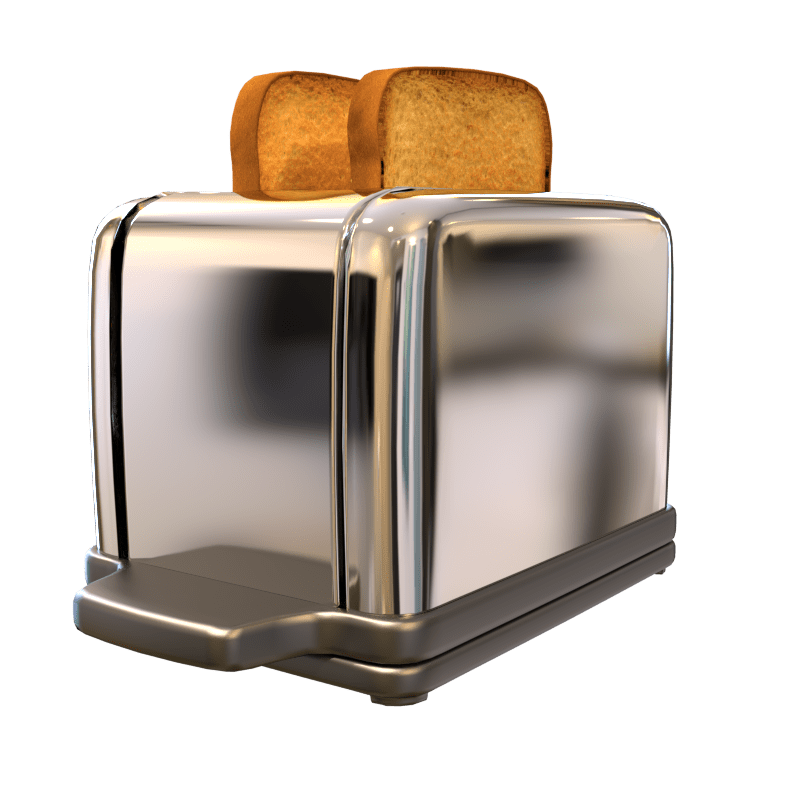} 
        \caption{Ground Truth} %
    \end{subfigure}%
    \vspace{-10pt}
    
    \caption{\textbf{Ablation of Environment Learner.} Eliminating $f_e$ results in bad rendering quality.}
  \label{fig:ablenvlearner}
    \end{minipage}
    \begin{minipage}{0.35\textwidth}
        \small
          \centering
            \setlength{\tabcolsep}{1.5mm}
           
            \begin{tabular}{c|cc}
            \toprule
                  & w/o $f_e$ & Ours \\
            \hline
            Toaster & 19.31 & 24.95 \\
            Materials & 27.37 & 29.58 \\
            \bottomrule
            \end{tabular}%
             \captionof{table}{\textbf{Ablation of Environment Learner.} Reported in PSNR.}
          \label{tab:ablenvlearnertab}%
    \end{minipage}
\end{figure}

\noindent \textbf{Environment Learner.} To validate the effectiveness of the environment learner $f_e$, we directly input $\bomega_r$ values into the tiny MLP $f_\psi$ as shown in~\cref{eq:cs_naive}, similar to~\cite{mobilenerf} and~\cite{nerf2mesh}, except they use viewing direction instead of reflective direction. Experimental results demonstrate that directly inputting $\bomega_r$ into the tiny MLP $f_\psi$ cannot learn the correct reflection appearance as a result of the limited network's capacity, as depicted in~\cref{fig:ablenvlearner}. Therefore, introducing the environment learner module improves the reconstruction quality without compromising the inference speed, as demonstrated in~\cref {tab:ablenvlearnertab}. 

\boldparagraph{Environment Feature Map} We can directly optimize an environment feature map instead of learning it through the environment learner module and baking it. However, as shown in~\cref{tab:ablenvmaptab}, the direct optimization of the environment feature map performs poorly when the map resolution is high. This is because, at higher resolutions, neighboring directions are mapped to distant grid points, leading to a non-smooth interpolation. As a result, the grid points cannot share global information and only contain its own local information. On the other hand, when the resolution is too low, the rendering quality is limited by the capacity of the environment feature map. In contrast, our method exhibits good rendering quality across different resolutions by learning the environment feature map via an MLP. 
Additionally, our environment feature maps enjoy the smoothness bias of the MLP, resulting in relatively smooth tensor maps. Consequently, when saving images in PNG format, our method demonstrates good compression performance. This further reduces the overhead of our caching.

\begin{table}[t]
\small
  \centering
    \setlength{\tabcolsep}{2.8mm}
    
    \begin{tabular}{cc|cccc}
    \toprule
          &       Resolutions& \multicolumn{2}{c}{Ours} & \multicolumn{2}{c}{Directly optimization} \\
          &       H*W& PSNR  & Cache(KB) & PSNR  & Cache(KB) \\
    \hline
    \multirow{2}[1]{*}{Toaster} & 180*360 & 24.86 & 49    & 24.16 & 91 \\
          & 900*1800 & 24.96 & 504   & 23.59 & 2460 \\
    \multirow{2}[1]{*}{Materials} & 180*360 & 29.55 & 63    & 29.58 & 114 \\
          & 900*1800 & 29.58 & 706   & 29.03 & 3000 \\
    \bottomrule
    \end{tabular}%
    \vspace{3pt}
\caption{\textbf{Ablation of Environment Feature Map.} Our method outperforms in both cache size and rendering quality.}
  \label{tab:ablenvmaptab}%
\end{table}%

\section{Conclusion}

Although our method successfully models the appearance of glossy objects, similar to Ref-NeRF~\cite{refnerf}, it still struggles with accurately modeling interreflections and non-distant illumination. Additionally, our method requires a certain level of quality in the initial mesh.

In summary, our method achieves real-time rendering on edge devices with low hardware budgets. Besides, we propose a novel approach for modeling view-dependent appearance and can optimize the appearance and geometry of glossy objects with high computational efficiency and low memory footprint. Furthermore, by effectively decoupling the scene's geometry, appearance, and environmental information, our method can perform simple scene editing tasks.

\section*{Acknowledgements}

This work is supported by NSFC under grant 62202418 and U21B2004. We thank Xiaoran Cao, Chenxi Tu, and Sicheng Li for their valuable discussions.

\clearpage  %

\bibliographystyle{splncs04}
\bibliography{main}

\clearpage
\vspace{5mm}
\appendix
\section*{\Large Appendix}
\renewcommand*{\thesection}{\Alph{section}}
\newcommand{\multiref}[2]{\cref{#1}--\ref{#2}}

In this \textbf{appendix}, we first present implementation details in~\cref{sec:implementation}. Next, we discuss the baselines implementations in \cref{sec:baselines}. Finally, we provide additional qualitative and quantitative results in~\cref{sec:experiment}.
    
\setcounter{footnote}{0}

\section{Implementation Details}
\label{sec:implementation}

In this section, we begin by describing how we obtain the initial mesh required for training (\cref{sec:initialmesh}). We then provide a detailed explanation of our network architecture (\cref{sec:network}) and our implementation details of training (\cref{sec:train}), rendering (\cref{sec:render}) and anti-aliasing (\cref{sec:AA}).

\subsection{Inital Mesh}
\label{sec:initialmesh}
\boldparagraph{NeRF Synthetic Dataset~\cite{nerf}} The NeRF Synthetic dataset consists of 100 training images and 200 testing images. Following the first stage of NeRF2Mesh~\cite{nerf2mesh}, we adopt a grid-based representation to extract our initial mesh. We refer to the first stage of NeRF2Mesh as ``NeRF2Mesh-S1'' in the following.

\boldparagraph{Shiny Blender Dataset~\cite{refnerf}} Similarly, the Shiny Blender dataset consists of 100 training images and 200 testing images. 
Nonetheless, the mesh extracted by NeRF2Mesh-S1 often results in low surface quality in areas that involve intricate view-dependent effects, as elaborated in~\cref{sec:reconstructionquality}. Therefore, we use the initial mesh extracted using Ref-Neus~\cite{refneus} as it is able to provide a faithful geometry reconstruction of glossy objects.

\boldparagraph{Real Captured Dataset~\cite{refnerf}} The Real Captured dataset includes three outdoor open scenes: gardenspheres, toycar, and sedan. 
In line with previous research, we select every eighth image from the input dataset as our test set. Similar to the Shiny Blender dataset, NeRF2Mesh-S1 performs poorly in modeling scenes with complex view-dependent appearances. Consequently, we employ Neuralangelo~\cite{neuralangelo} to extract the initial mesh for the Real Captured scenes. However, due to the poor reconstruction quality of Neuralangelo~\cite{neuralangelo} on the gardenspheres scene (as detailed in~\cref{sec:reconstructionquality}), we still use the inaccurate mesh extracted by NeRF2Mesh-S1 as our geometric prior in gardenspheres scene. 

\subsection{Network Architecture}
\label{sec:network}
\boldparagraph{Geometry Learner} The geometry learner module includes $g_{{\theta}_{\bv}}$ and $g_{\theta_\bn}$. Both networks are composed of multi-resolution hash tables~\cite{instantngp} and a small MLP. The hash table structures of both networks are identical. For object-centric datasets, we adopt hash tables with 16 layers, a maximum table length of $2^{19}$, a minimum resolution of 16, and a maximum resolution of 512. For the unbounded Real Captured dataset, we increase the maximum resolution and use a maximum table length of  $2^{20}$.

\boldparagraph{Environment Learner} The environment learner module is a four-layer MLP with a width of 256. The output of the environment learner module is an $M$-dimensional environment feature. Through experimental validation, it has been determined that $M=3$ achieves high reconstruction quality without imposing significant overhead on caching.

\boldparagraph{Color Net} The color net is divided into two parts, $f_\phi$ and $f_\psi$. $f_\phi$ consists of multi-resolution hash tables~\cite{instantngp} with the same structure as $g_{\theta_\bn}$ and a single-layer MLP. $f_\psi$ is a 2-layer MLP with a width of 64. During the mobile rendering stage, we query $f_\psi$ as the only network, as all other information can be baked into texture images. 

\subsection{Training Implementation}
\label{sec:train}
\boldparagraph{Training Details} Our method utilizes the Adam~\cite{adam} optimizer and employs the cosine annealing learning rate adjustment strategy~\cite{cos} with a period of 2 and a period step of 2. We train our model for 250 epochs on each scene. With a batch size of 1, we render only one image per iteration and calculate the loss to optimize our network. During training, we set $\lambda_{\bc}$ to 1, $\lambda_{\bc_d}$ to 0.001, $\lambda_{SSIM}$ to 3, $\lambda_{mask}$ to 100, $\lambda_{\Delta\bn_\bv}$ to 0.1, $\lambda_{max}$ to 0.00001. Given the coarse initial mesh, we scale normals and vertex offsets with $w=\min(e/e_{warm},1)$, where $e$ represents the current epoch and $e_{warm}$ is the warm-up epoch number to stabilize training process.

\boldparagraph{Real Captured Dataset} For the Real Captured dataset, due to the specular appearance often occurring on foreground objects in outdoor scenes, we use different environment learners for foreground and background to learn the view-dependent appearance separately in order to avoid interference from the background.

\subsection{Rendering Implementation}
\label{sec:render}
\boldparagraph{Texture Images} During the mobile rendering stage, to accelerate the inference process, we bake the $\bc_d$ and $\bff_s$ obtained from the $f_\phi$ as two texture images with a resolution of 4096 $\times$ 4096. These images are saved in JPG format for compression purposes. Similarly, we bake the normal information as a texture image with the same resolution. However, to ensure geometric accuracy, we select the PNG format as a lossless compression method to store our normal texture image. Besides, accurate estimation of reflection directions is paramount for the view-dependent appearance modeling of highly reflective surfaces. Therefore, for the Shiny Blender dataset, we store the normal information in two PNG images. Specifically, we first bake the $\bn_\bp$ in one PNG image, denoted as $\bn'_\bp$. Then, we calculate the residuals of the $\bn'_\bp$ relative to the $\bn_\bp$ and save these residuals in another PNG image. It is important to note that since the residuals are typically quite small, we divide the residuals by their average order of magnitude before saving the residual image to cut its compression loss.

\boldparagraph{Environment Feature Map} Querying the environment learner during the rendering process to obtain the environment feature $\bff_e$ significantly reduces our rendering speed. To address this, we need to bake the environment learner into an environment feature map. This allows us to directly query the environment feature map using the reflection direction $\bomega_r$ during rendering, greatly accelerating the inference speed. Additionally, to reduce caching overhead, we convert the three-dimensional reflection direction into two-dimensional polar coordinates, as shown in~\cref{eq:dir2polar}:

\begin{equation}
    \left\{\begin{matrix}\theta = arccos(\frac{z}{\sqrt{x^2+y^2+z^2} } )\\\phi = arccos(\frac{x}{\sqrt{x^2+y^2}})\ast \frac{y}{|y|}\end{matrix}\right.
\label{eq:dir2polar}
\end{equation}
where x, y, and z represent the components of $\bomega_r$ in three different directions. To obtain the environment feature map, we uniformly sample a series of grids on the environment feature map. We then transform each grid to corresponding three-dimensional $\bomega_r$ (see~\cref{eq:polar2dir}) and input them into the environment learner to query $\bff_e$:
\begin{equation}
    \left\{\begin{matrix}
    z = cos(\theta)
    \\x = \sqrt{1-z^2}\ast  cos(\phi)
    \\y = \sqrt{1-z^2-x^2}\ast \frac{\phi}{|\phi|} 
\end{matrix}\right.
\label{eq:polar2dir}
\end{equation}
During rendering, for each direction, we can convert it into polar coordinates and use bilinear interpolation to retrieve $\bff_e$ on the environment feature map. High-quality environment features are crucial for modeling the view-dependent appearance of highly reflective surfaces since they capture the ambient lighting information of the scene. Hence, akin to normal texture images, we employ two PNG images to store the environment feature map, enabling us to obtain high-quality environment features during the rendering stage.

\subsection{Anti-aliasing Implementation}
\label{sec:AA}
\boldparagraph{Nvdiffrast Anti-aliasing} During training stage, we utilize Nvdiffrast~\cite{difras} as our differentiable rasterizer. Nvdiffrast relies on anti-aliasing to backpropagate gradients, by calculating an approximate integral over a pixel based on the exact location of relevant edges and the point-sampled colors at pixel centers. However, its anti-aliasing operation is not supported on mobile devices. Therefore, the results reported in Ours (Mobile) exclude the anti-aliasing from Nvdiffrast. On the other hand, to ensure a fair comparison with baselines, version Ours includes Nvdiffrast's anti-aliasing, following the setting of~\cite{nerf2mesh}.

\boldparagraph{Upsampling Anti-aliasing} For object-level scenes, in order to eliminate the jagged edges, we follow the anti-aliasing strategy adopted by NeRF2Mesh~\cite{nerf2mesh} and MobileNeRF~\cite{mobilenerf}. Specifically, we render the image at an upsampled resolution ($2\times 2$) and downsample it to the target resolution for anti-aliasing. It's important to note that this technique is integrated into our implementation for mobile devices. Consequently, all the reported results for object-level scenes in our study include this anti-aliasing implementation, same as the baselines~\cite{nerf2mesh,mobilenerf}.

\section{Baselines}
\label{sec:baselines}
\boldparagraph{Ref-NeRF~\cite{refnerf}} We introduce Ref-NeRF~\cite{refnerf} as the state-of-the-art approach for rendering glossy objects. Unfortunately, our attempt to reproduce the results from the original paper using its official implementation\footnote{https://github.com/google-research/multinerf} is unsuccessful. Therefore, we directly reference the quantitative and qualitative results from the original paper for comparison, in~\cref{tab:renderingquality} and~\cref{fig:gardensphere} of our main paper, respectively. 

\boldparagraph{3DGS~\cite{3DGS}} We introduce 3DGS~\cite{3DGS} as the state-of-the-art work that enables real-time rendering on GPU. We utilize the official implementation\footnote{https://github.com/graphdeco-inria/gaussian-splatting}of 3DGS~\cite{3DGS} to get the results for the Real Captured dataset. For the NeRF Synthetic and Shiny Blender dataset, we quote the results from the GaussianShader~\cite{gaussianshader} paper.

\boldparagraph{NvDiffRec~\cite{NvDiffRec}} We introduce NvDiffRec~\cite{NvDiffRec} as a representative work for physically-based rendering methods. NvDiffRec~\cite{NvDiffRec} requires a mask as supervision, limiting its applicability to objective-level scenes. Consequently, we are unable to report its results on Real Captured scenes in~\cref{tab:renderingquality} of our main text. We use the official NvDiffRec~\cite{NvDiffRec} implementation\footnote{https://github.com/NVlabs/nvdiffrec} to reproduce their results on the Shiny Blender dataset. It is worth noting that NvDiffRec~\cite{NvDiffRec} has different configuration files for different scenes in the NeRF Synthetic dataset. For the Shiny Blender dataset, we use the configuration for the materials scene since it contains the richest view-dependent information in the NeRF Synthetic dataset. For the NeRF Synthetic dataset, we reference the results from the original paper of NvDiffRec~\cite{NvDiffRec}.

\boldparagraph{MobileNeRF~\cite{mobilenerf}} We introduce MobileNeRF~\cite{mobilenerf} as a representative work that enables real-time rendering on mobile devices. We utilize the official implementation\footnote{https://github.com/google-research/jax3d/tree/main/jax3d/projects/mobilenerf} of MobileNeRF~\cite{mobilenerf} to obtain results for the Real Captured dataset and Shiny Blender dataset. For the NeRF Synthetic dataset, we reference the results from the original paper of MobileNeRF~\cite{mobilenerf}.

\boldparagraph{NeRF2Mesh~\cite{nerf2mesh}} We introduce NeRF2Mesh~\cite{nerf2mesh} as another representative work that enables real-time rendering on mobile devices. We utilize the official implementation\footnote{https://github.com/ashawkey/nerf2mesh} of NeRF2Mesh~\cite{nerf2mesh} provided by its authors to get results for the Real Captured dataset and Shiny Blender dataset. For the NeRF Synthetic dataset, we reference the results from the original paper of NeRF2Mesh~\cite{nerf2mesh}. We use the $w/ \mathcal{L}_{smooth}$ version of NeRF2Mesh~\cite{nerf2mesh} with mask supervision which is the default setting in its implementation. In addition, since NeRF2Mesh~\cite{nerf2mesh} also requires an initial mesh in its second stage, for a fair comparison, we provide the same initial mesh to NeRF2Mesh~\cite{nerf2mesh} for both the Shiny Blender dataset and the Real Captured dataset.

\section{Additional Experimental Results}
\label{sec:experiment}
In this section, we begin by quantitatively and qualitatively comparing the reconstruction quality of our method with the baseline methods~\cite{nerf2mesh,mobilenerf,refnerf,NvDiffRec,3DGS} (\cref{sec:reconstructionquality}). Then, we provide a detailed analysis of the rendering efficiency of our method on various devices (\cref{sec:renderingefficiency}). Additionally, we supplement the study with two ablation experiments to further demonstrate the effectiveness of our method (\cref{sec:abl}). Finally, we showcase the performance of our method in several downstream applications (\cref{sec:sceneediting}).
\subsection{Reconstruction Quality}
\label{sec:reconstructionquality}

\begin{figure}[!htbp] 
  \centering
   \includegraphics[width=1\linewidth]{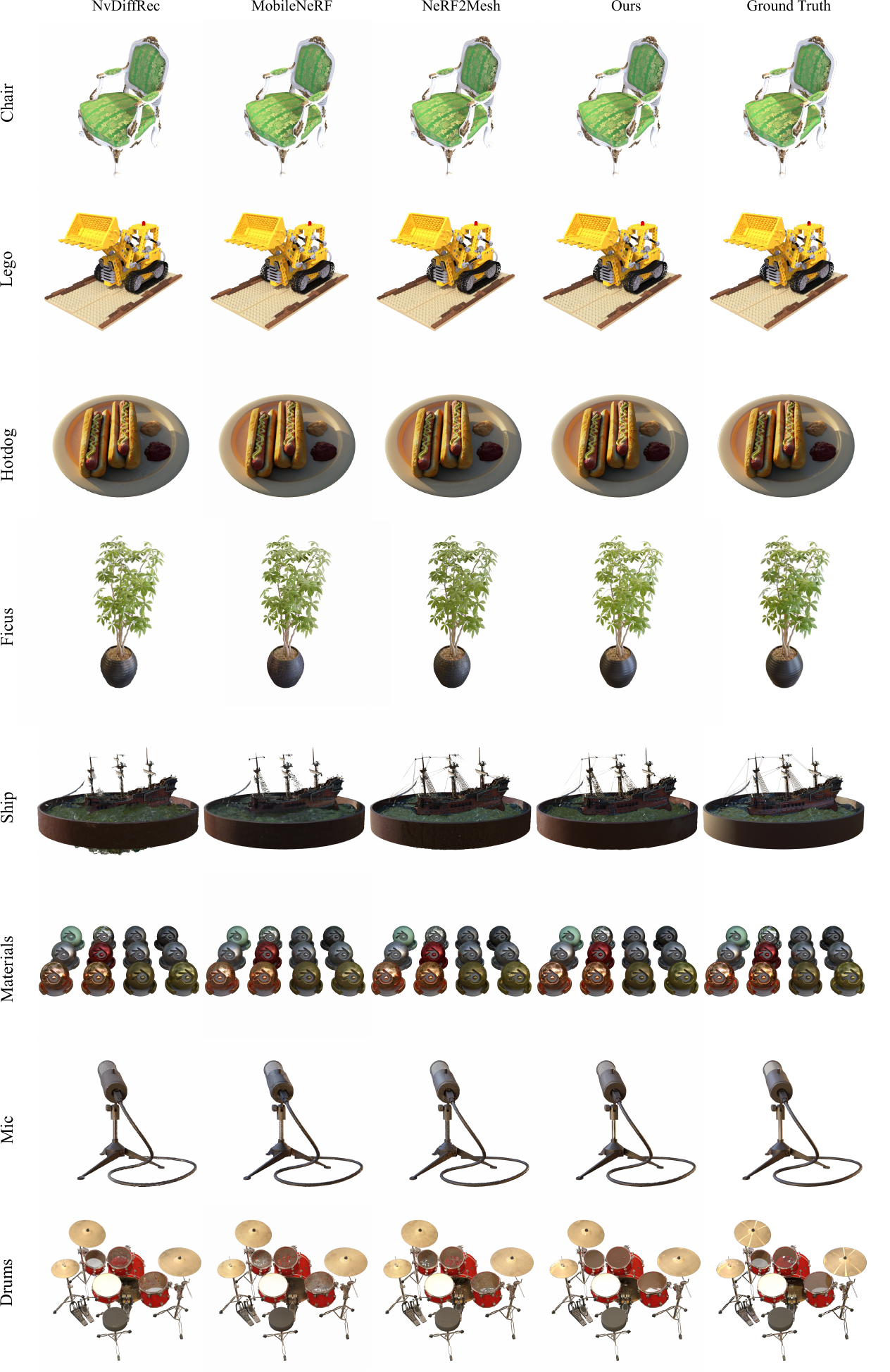}
   \caption{\textbf{Qualitative Comparison.} We provide visualization results for our method and the baseline methods~\cite{mobilenerf,NvDiffRec,nerf2mesh} on the NeRF Synthetic dataset.}
   \label{fig:visualization}
\end{figure}

\begin{figure}[!htbp] 
  \centering
   \includegraphics[width=1.0\linewidth]{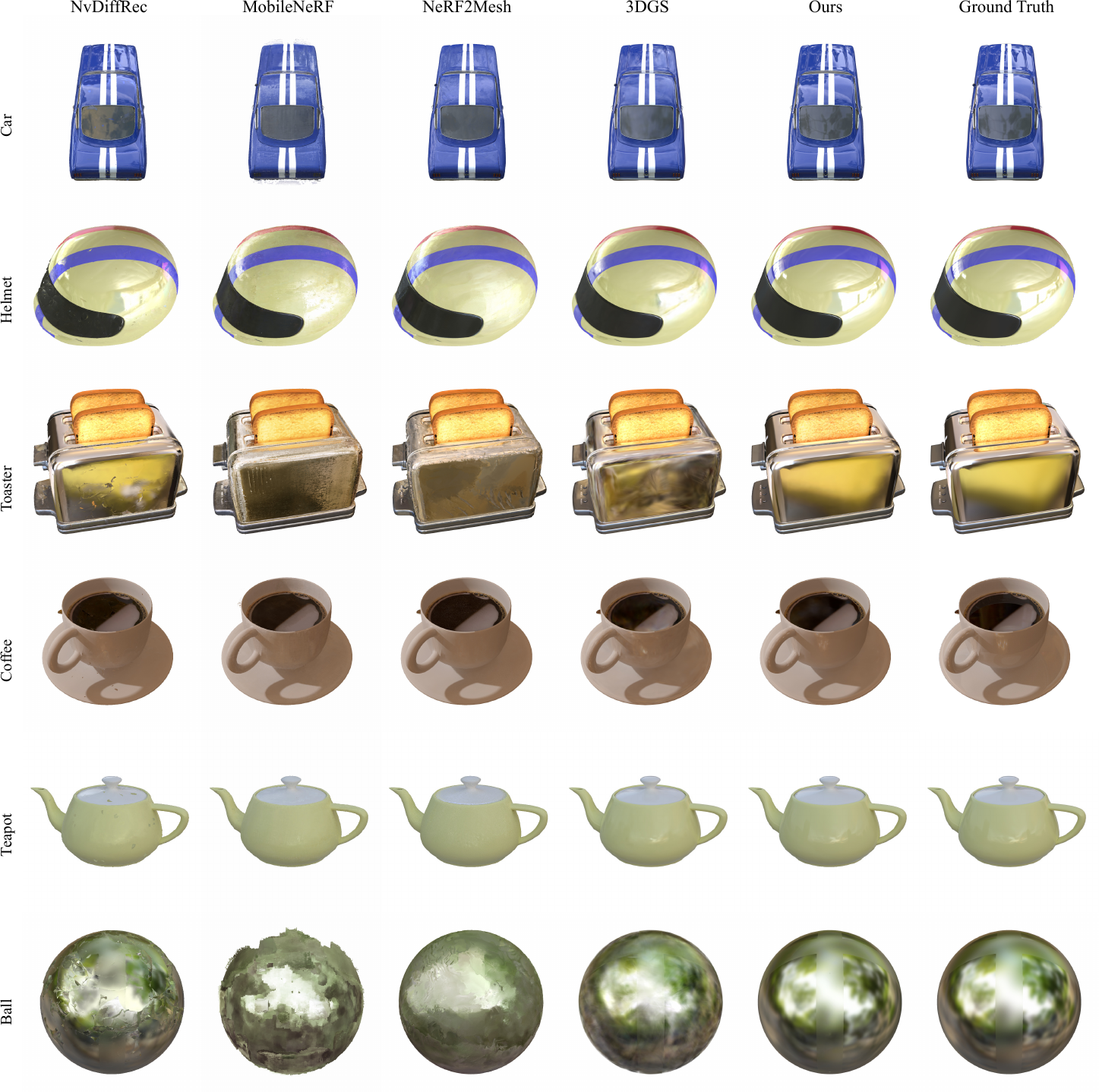}
   \caption{\textbf{Qualitative Comparison.} We provide visualization results for our method and the baseline methods~\cite{mobilenerf,NvDiffRec,nerf2mesh,3DGS} on the Shiny Blender dataset.}
   \label{fig:shinyquality}
\end{figure}

\begin{figure}[h]
  \centering
   \includegraphics[width=1\linewidth]{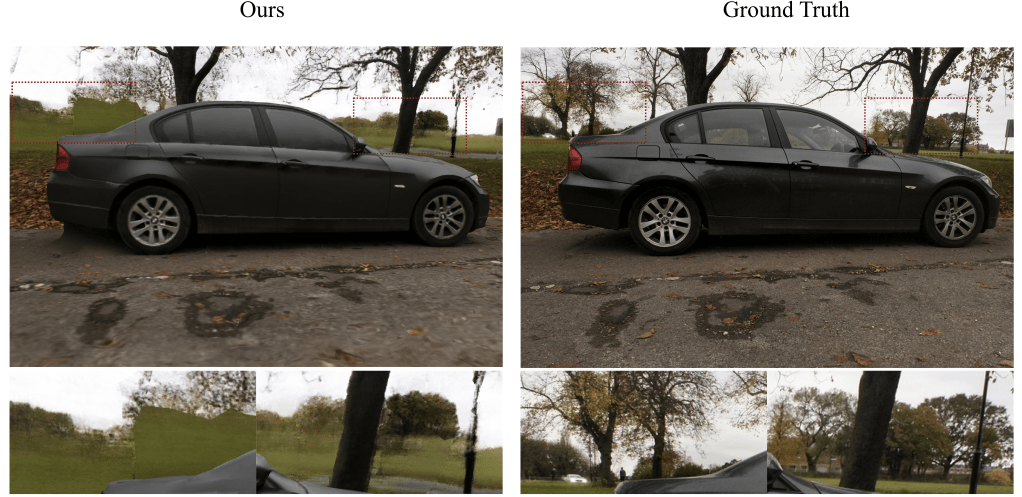}
   \caption{\textbf{Rendering Result in Real Captured Dataset.} Our PSNR metric is primarily limited by the background.}
   \label{fig:backgroundendering}
\end{figure}

\begin{figure}[h]
  \centering
   \includegraphics[width=1.0\linewidth]{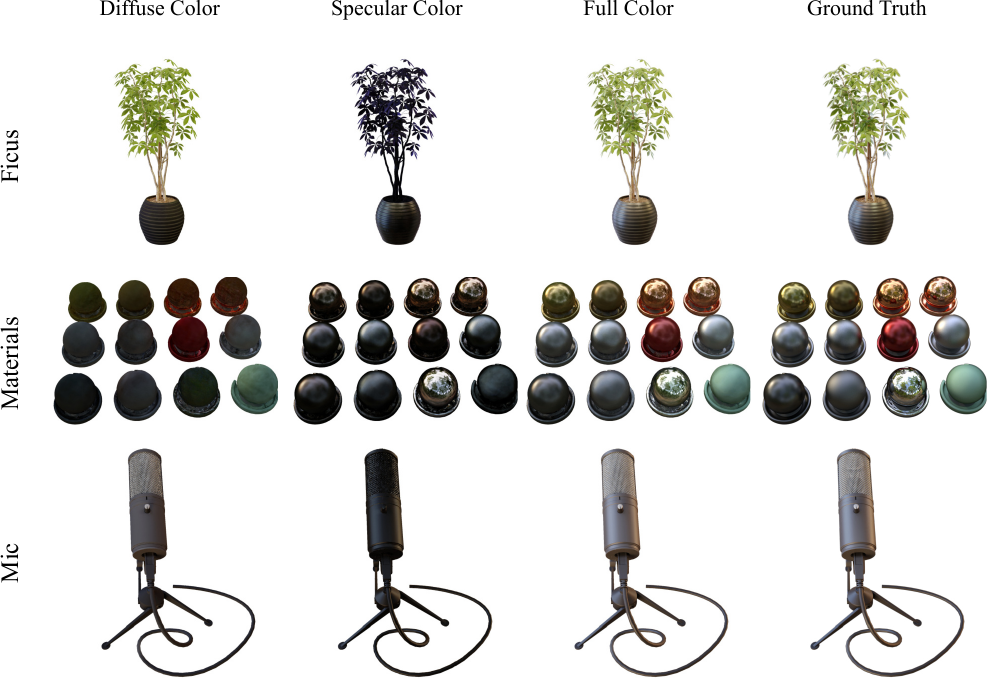}
   \caption{\textbf{Color Decoupling.} Our method is able to effectively decouple the diffuse color and specular color.}
   \label{fig:difspe}
\end{figure}

\begin{figure}[t] %
    \begin{subfigure}[t]{0.5\linewidth} %
        \centering
        \includegraphics[width=1\linewidth]{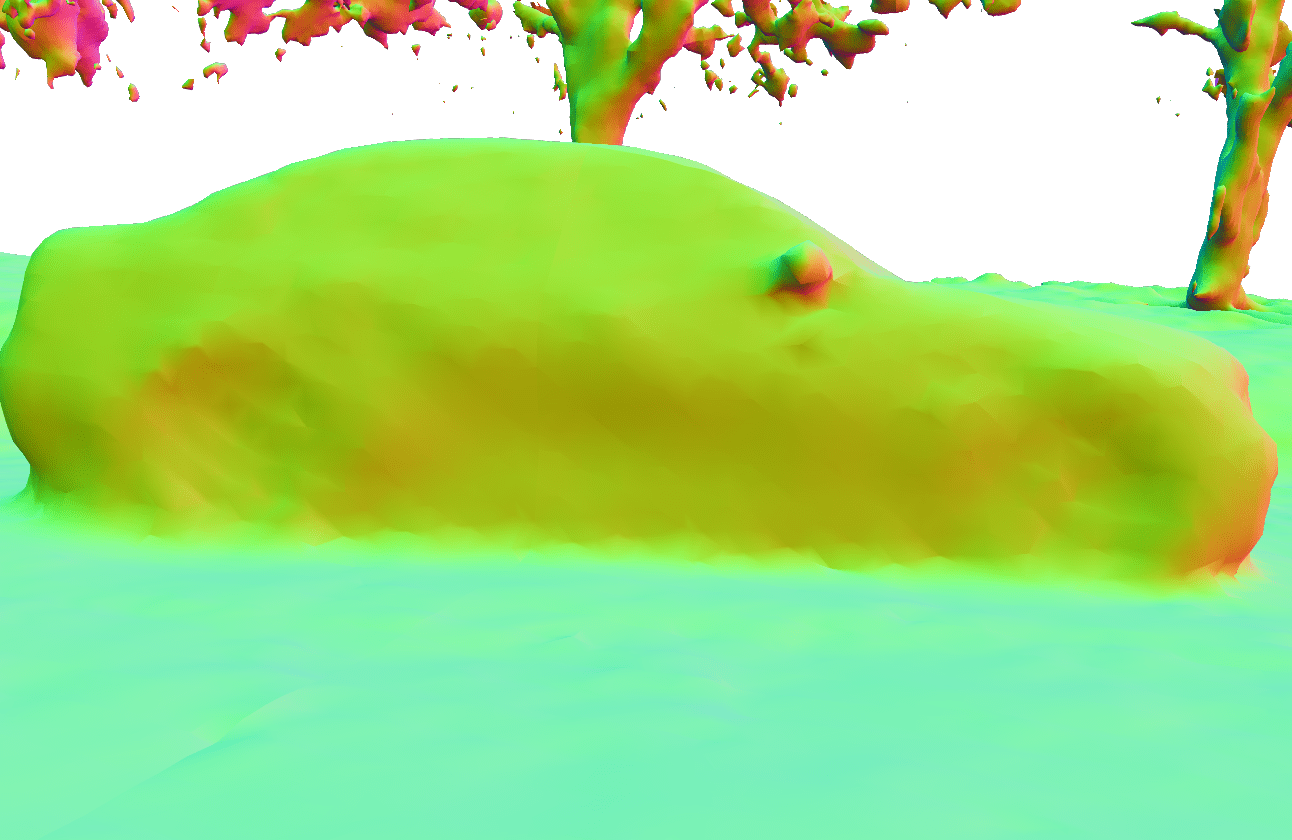} 
        \caption{Normal.} %
    \end{subfigure}%
    \begin{subfigure}[t]{0.5\linewidth} %
        \centering
        \includegraphics[width=1\linewidth]{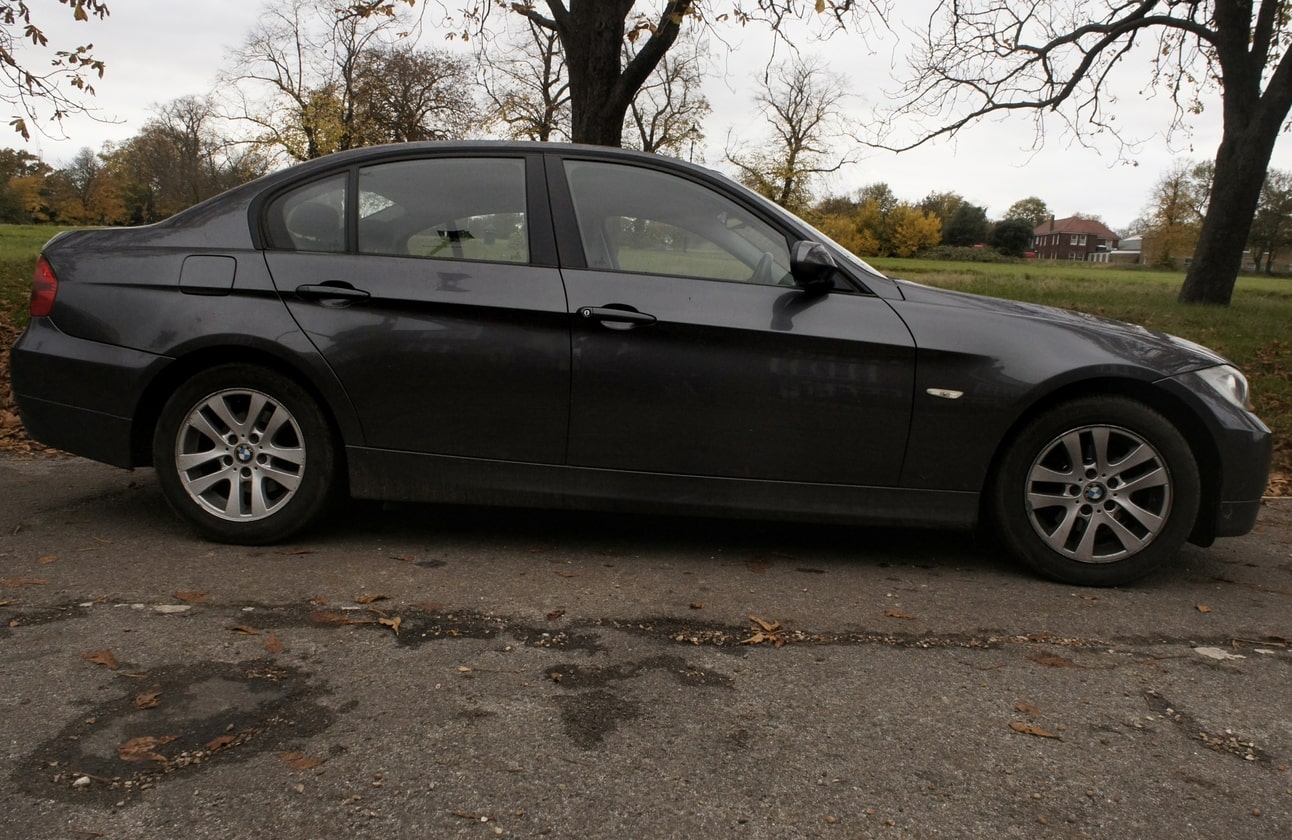} 
        \caption{Ground Truth.} %
    \end{subfigure}%
    \caption{\textbf{Mesh Extracted by Neuralangelo~\cite{neuralangelo}.} The left figure illustrates the normals obtained by rendering the mesh extracted by Neuralangelo~\cite{neuralangelo} from the corresponding viewpoints of the ground truth. While Neuralangelo~\cite{neuralangelo} can reconstruct foreground objects well, it still struggles to accurately model background details.}
  \label{fig:neuralangeloiniitalmesh}
\end{figure}

\boldparagraph{Rendering Quality} For more detailed quantitative metric comparisons, please refer to~\cref{tab:psnrnerf,tab:ssimnerf,tab:lpipsnerf,tab:psnrshiny,tab:ssimshiny,tab:lpipsshiny,tab:psnrreal,tab:ssimreal,tab:lpipsreal}. We also demonstrate a qualitative comparison between our method and the mesh-based baselines~\cite{mobilenerf,NvDiffRec,nerf2mesh} on the NeRF Synthetic dataset, as shown in~\cref{fig:visualization}. We also compare with all real-time rendering baselines~\cite{3DGS,mobilenerf,nerf2mesh,NvDiffRec} on the Shiny Blender dataset, as shown in~\cref{fig:shinyquality}. Our method achieves the best rendering quality among real-time rendering approaches~\cite{mobilenerf,NvDiffRec,nerf2mesh,3DGS} for the Shiny Blender dataset where our rendering quality even surpasses the non-real-time state-of-the-art rendering work~\cite{refnerf}. However, our method does not perform well in terms of PSNR on Real Captured scenes. This is because our method is limited by the initial mesh (\cref{fig:neuralangeloiniitalmesh}) and does not reconstruct the appearance of the background well, as shown in~\cref{fig:backgroundendering}. This has a significant impact on our PSNR, as the background often occupies a large portion of the image. On the other hand, our method excels in recovering the appearance of foreground objects, resulting in better subjective quality (SSIM~\cite{ssim} and LPIPS~\cite{lpips}) compared to mesh-based methods. In addition, our method effectively decouples the diffuse appearance and specular appearance, as demonstrated in~\cref{fig:difspe}.

\begin{figure}[h] %
    \begin{subfigure}[t]{0.25\linewidth} %
        \centering
        \includegraphics[width=1\linewidth]{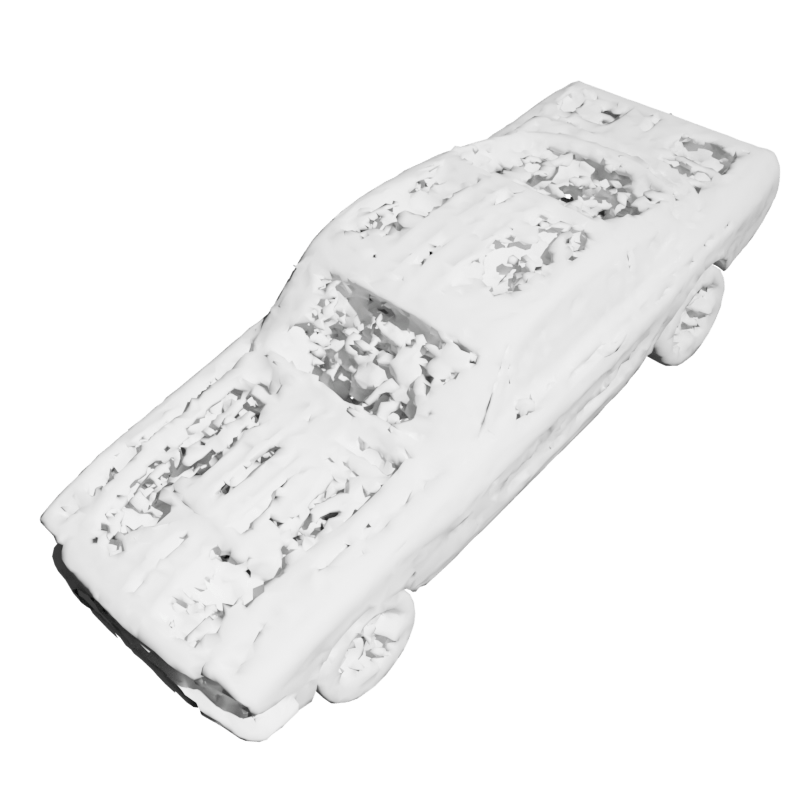} 
        \caption{Car.} %
    \end{subfigure}%
    \begin{subfigure}[t]{0.25\linewidth} %
        \centering
        \includegraphics[width=1\linewidth]{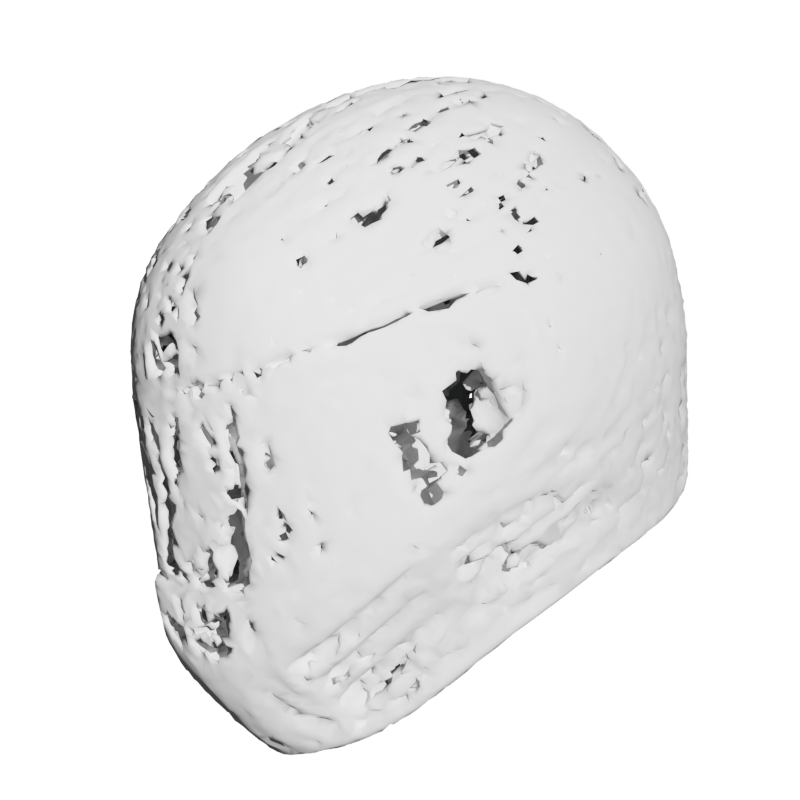} 
        \caption{Helmet.} %
    \end{subfigure}%
    \begin{subfigure}[t]{0.25\linewidth} %
        \centering
        \includegraphics[width=1\linewidth]{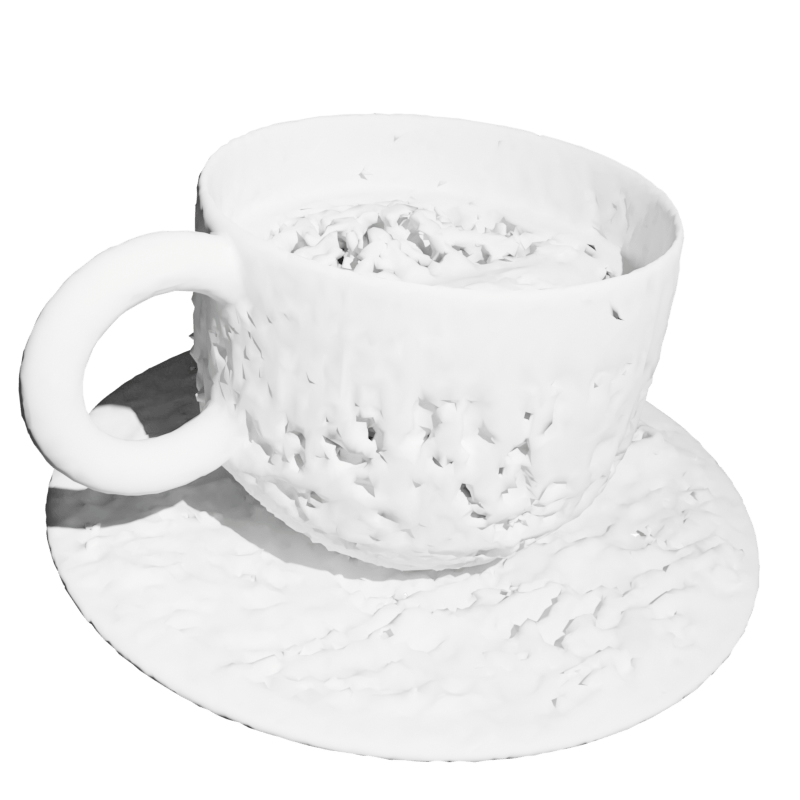} 
        \caption{Coffee.} %
    \end{subfigure}%
    \begin{subfigure}[t]{0.25\linewidth} %
        \centering
        \includegraphics[width=1\linewidth]{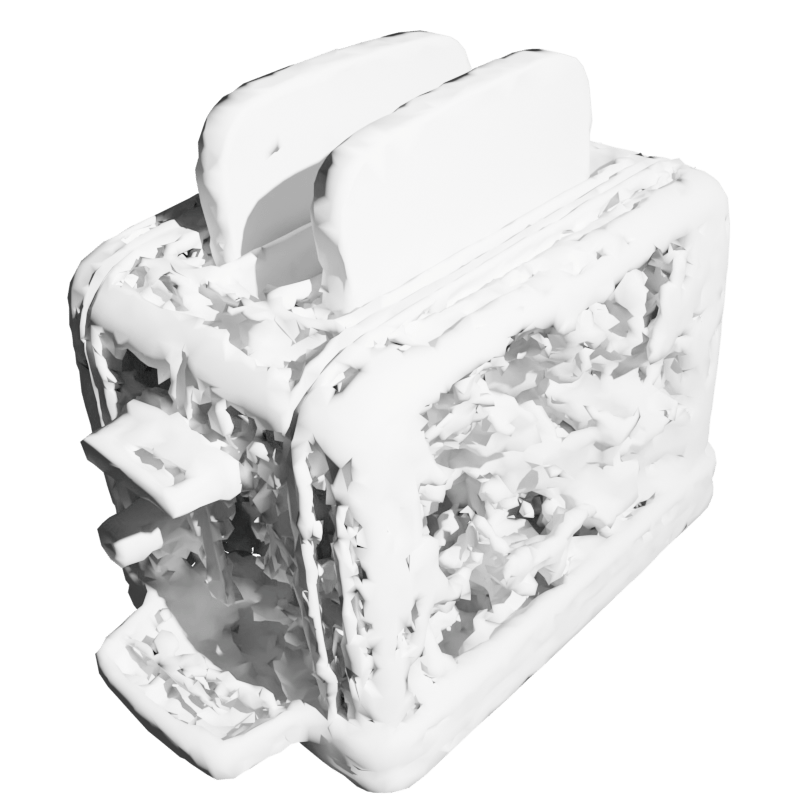} 
        \caption{Toaster.} %
    \end{subfigure}%
    \caption{\textbf{Mesh Extracted by NeRF2Mesh-S1.} NeRF2Mesh-S1 struggles to reconstruct the geometry of glossy objects.}
  \label{fig:nerf2meshinitialmesh}
\end{figure}

\begin{figure}[t] %
    \begin{subfigure}[t]{0.5\linewidth} %
        \centering
        \includegraphics[width=1\linewidth]{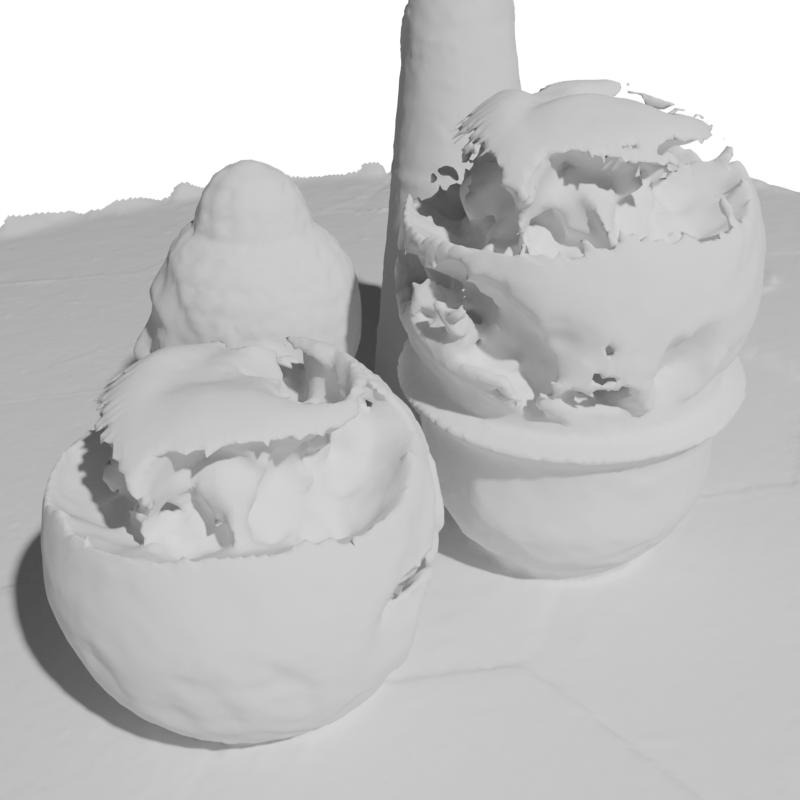} 
        \caption{Neuralangelo.} %
    \end{subfigure}%
    \begin{subfigure}[t]{0.5\linewidth} %
        \centering
        \includegraphics[width=1\linewidth]{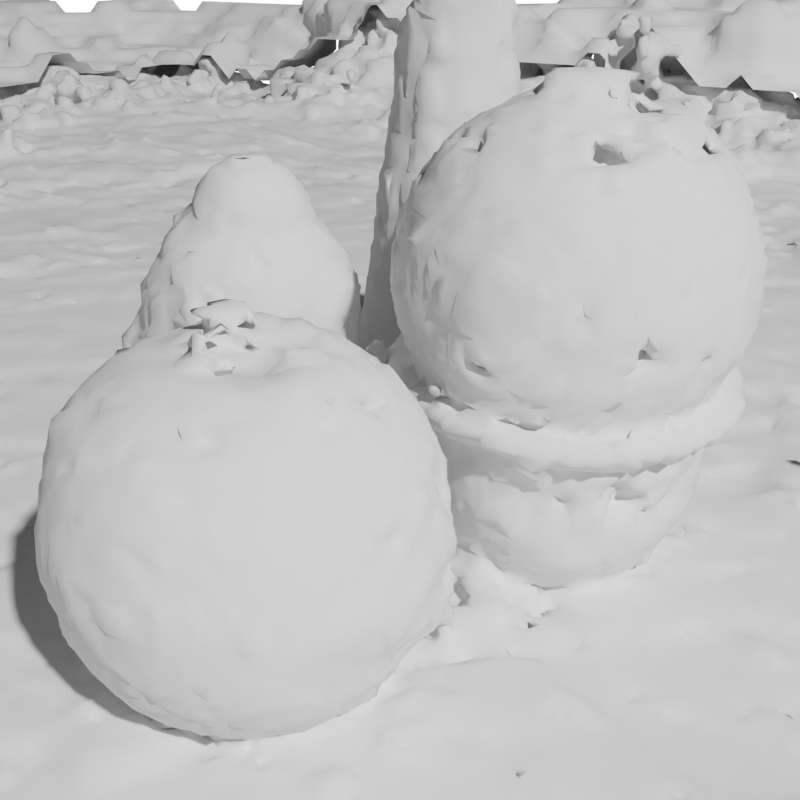} 
        \caption{NeRF2Mesh-S1.} %
    \end{subfigure}%
    \caption{\textbf{Mesh for Gardenspheres.} Neuralangelo~\cite{neuralangelo} fails to accurately reconstruct the geometry of the gardenspheres scene.}
  \label{fig:gardenspheremesh}
\end{figure}

\boldparagraph{Initial Surface Quality} As mentioned in~\cref{sec:initialmesh}, the mesh extracted by NeRF2Mesh-S1 for the Shiny Blender dataset can not depict the accurate geometry, as shown in~\cref{fig:nerf2meshinitialmesh}. Therefore, we utilize the official implementation\footnote{https://github.com/EnVision-Research/Ref-NeuS} provided by the authors of Ref-Neus~\cite{refneus} to extract the initial mesh on the Shiny Blender dataset. Besides, we use the official implementation\footnote{https://github.com/nvlabs/neuralangelo} of Neuralangelo~\cite{neuralangelo} to extract our initial mesh for the sedan and toycar. Although it does not model the background well, Neuralangelo~\cite{neuralangelo} provides accurate estimates for the foreground objects, as shown in~\cref{fig:neuralangeloiniitalmesh}. Due to the presence of moving pedestrians, distant trees with intricate details, houses, and vehicles, the background in the sedan scene poses a particularly challenging task for mesh-based modeling. However, modeling the background of such scenes is not the primary focus of our work. In the meantime, the initial mesh extracted by Neuralangelo~\cite{neuralangelo} for the gardenspheres scene has poor geometry. Therefore, we utilize the flawed initial mesh obtained by NeRF2Mesh-S1~\cite{nerf2mesh} as our geometric initialization, as shown in~\cref{fig:gardenspheremesh}.
\begin{table}[!tbp]
  \centering
    \small
    \setlength{\tabcolsep}{1mm}
    \begin{tabular}{c|ccccccccc}
    \toprule
          & Chair & Drums & Ficus & Hotdog & Lego  & Mats. & Mic   & Ship  & Mean \\
    \hline
    Ref-NeRF~\cite{refnerf} & \cellcolor[rgb]{ 1,  .8,  .8}35.83  & \cellcolor[rgb]{ .973,  .796,  .678}25.79  & \cellcolor[rgb]{ .973,  .796,  .678}33.91  & \cellcolor[rgb]{ 1,  .8,  .8}37.72  & \cellcolor[rgb]{ 1,  .8,  .8}36.25  & \cellcolor[rgb]{ 1,  .8,  .8}35.41  & \cellcolor[rgb]{ 1,  .8,  .8}36.76  & \cellcolor[rgb]{ .973,  .796,  .678}30.28  & \cellcolor[rgb]{ 1,  .8,  .8}33.99  \\
    \hline
    3DGS~\cite{3DGS} & \cellcolor[rgb]{ .973,  .796,  .678}35.82 & \cellcolor[rgb]{ 1,  .8,  .8}26.17  & \cellcolor[rgb]{ 1,  .8,  .8}34.83  & \cellcolor[rgb]{ .973,  .796,  .678}37.67  & \cellcolor[rgb]{ .973,  .796,  .678}35.69  & \cellcolor[rgb]{ .973,  .796,  .678}30.00  & \cellcolor[rgb]{ .973,  .796,  .678}35.34  & \cellcolor[rgb]{ 1,  .8,  .8}30.87  & \cellcolor[rgb]{ .973,  .796,  .678}33.30  \\
    NvDiffRec~\cite{NvDiffRec} & 31.60  & 24.10  & 30.88  & 33.04  & 29.14  & 26.74  & 30.78  & 26.12  & 29.05  \\
    MobileNeRF~\cite{mobilenerf} & \cellcolor[rgb]{ 1,  .949,  .8}34.09  & 25.02  & 30.20  & \cellcolor[rgb]{ 1,  .949,  .8}35.46  & \cellcolor[rgb]{ 1,  .949,  .8}34.18  & 26.72  & 32.48  & \cellcolor[rgb]{ 1,  .949,  .8}29.06  & 30.90  \\
    NeRF2Mesh~\cite{nerf2mesh} & 31.93  & 24.80  & 29.81  & 34.11  & 32.07  & 25.45  & 31.25  & 28.69  & 29.76  \\
    Ours  & 34.02  & \cellcolor[rgb]{ 1,  .949,  .8}25.07  & \cellcolor[rgb]{ 1,  .949,  .8}31.54  & 34.89  & 32.61  & \cellcolor[rgb]{ 1,  .949,  .8}29.58  &\cellcolor[rgb]{ 1,  .949,  .8}32.84  & 27.73  & \cellcolor[rgb]{ 1,  .949,  .8}31.04  \\
    \hline
    Ours (Mobile) & 33.94 & 25.00 & 31.20 & 34.66 & 32.21 & 29.45 & 32.67 & 27.56 & 30.84 \\
    \bottomrule
    \end{tabular}%
    \vspace{3pt}
    \caption{\textbf{PSNR$\uparrow$ on NeRF Synthetic Dataset.} Red represents the optimal, orange represents the second best, and yellow represents the third.}
  \label{tab:psnrnerf}%
\end{table}%

\begin{table}[!tbp]
  \centering
    \small
    \setlength{\tabcolsep}{1mm}
    \begin{tabular}{c|ccccccccc}
    \toprule
          & Chair & Drums & Ficus & Hotdog & Lego  & Mats. & Mic   & Ship  & Mean \\
    \hline
    Ref-NeRF~\cite{refnerf} & \cellcolor[rgb]{ .973,  .796,  .678}0.984  & 0.937  & \cellcolor[rgb]{ .973,  .796,  .678}0.983  & \cellcolor[rgb]{ .973,  .796,  .678}0.984  & \cellcolor[rgb]{ .973,  .796,  .678}0.981  &\cellcolor[rgb]{ 1,  .8,  .8}0.983  & \cellcolor[rgb]{ 1,  .8,  .8}0.992  & \cellcolor[rgb]{ .973,  .796,  .678}0.880  & \cellcolor[rgb]{ .973,  .796,  .678}0.966  \\
    \hline
    3DGS~\cite{3DGS} & \cellcolor[rgb]{ 1,  .8,  .8}0.987 & \cellcolor[rgb]{ 1,  .8,  .8}0.954  & \cellcolor[rgb]{ 1,  .8,  .8}0.987  &\cellcolor[rgb]{ 1,  .8,  .8}0.985  & \cellcolor[rgb]{ 1,  .8,  .8}0.983  & 0.960  & \cellcolor[rgb]{ .973,  .796,  .678}0.991  & \cellcolor[rgb]{ 1,  .8,  .8}0.907  & \cellcolor[rgb]{ 1,  .8,  .8}0.969  \\
    NvDiffRec~\cite{NvDiffRec} & 0.969 & 0.916  & 0.970  & 0.973  & 0.949  & 0.923  & 0.977  & 0.833  & 0.939  \\
    MobileNeRF~\cite{mobilenerf} & \cellcolor[rgb]{ 1,  .949,  .8}0.978 & 0.927  & 0.965  & 0.973  & \cellcolor[rgb]{ 1,  .949,  .8}0.975  & 0.913  & 0.979  & 0.867  & 0.947  \\
    NeRF2Mesh~\cite{nerf2mesh} & 0.964  & 0.927  & 0.967  & 0.970  & 0.957  & 0.896  & 0.974  & 0.865  & 0.940  \\
    Ours  & \cellcolor[rgb]{ .973,  .796,  .678}0.984  & \cellcolor[rgb]{ .973,  .796,  .678}0.947  & \cellcolor[rgb]{ 1,  .949,  .8}0.980  & \cellcolor[rgb]{ 1,  .949,  .8}0.982  & 0.973  & \cellcolor[rgb]{ .973,  .796,  .678}0.962  & \cellcolor[rgb]{ 1,  .949,  .8}0.987  & \cellcolor[rgb]{ 1,  .949,  .8}0.874  &\cellcolor[rgb]{ 1,  .949,  .8}0.961  \\
    \hline
    Ours (Mobile) & \cellcolor[rgb]{ .973,  .796,  .678}0.984 & \cellcolor[rgb]{ 1,  .949,  .8}0.946  & 0.979  & 0.979  & 0.969  & \cellcolor[rgb]{ 1,  .949,  .8}0.961  & 0.986  & 0.871  & 0.959  \\
    \bottomrule
    \end{tabular}%
    \vspace{3pt}
    \caption{\textbf{SSIM$\uparrow$ on NeRF Synthetic Dataset.} Red represents the optimal, orange represents the second best, and yellow represents the third.}
  \label{tab:ssimnerf}%
\end{table}%

\begin{table}[!tbp]
  \centering
  \small
      \setlength{\tabcolsep}{1mm}
    \begin{tabular}{c|ccccccccc}
    \toprule
          & Chair & Drums & Ficus & Hotdog & Lego  & Mats. & Mic   & Ship  & Mean \\
    \hline
    Ref-NeRF~\cite{refnerf} & \cellcolor[rgb]{ .973,  .796,  .678}0.017  & 0.059  & \cellcolor[rgb]{ .973,  .796,  .678}0.019  & \cellcolor[rgb]{ .973,  .796,  .678}0.022  & \cellcolor[rgb]{ .973,  .796,  .678}0.018  & \cellcolor[rgb]{ 1,  .8,  .8}0.022  & \cellcolor[rgb]{ .973,  .796,  .678}0.007  & 0.139  & \cellcolor[rgb]{ .973,  .796,  .678}0.038  \\
    \hline
    3DGS~\cite{3DGS} & \cellcolor[rgb]{ 1,  .8,  .8}0.012  & \cellcolor[rgb]{ 1,  .8,  .8}0.037  & \cellcolor[rgb]{ 1,  .8,  .8}0.012  & \cellcolor[rgb]{ 1,  .8,  .8}0.020  & \cellcolor[rgb]{ 1,  .8,  .8}0.016  & \cellcolor[rgb]{ .973,  .796,  .678}0.034  & \cellcolor[rgb]{ 1,  .8,  .8}0.006  & \cellcolor[rgb]{ 1,  .8,  .8}0.106  & \cellcolor[rgb]{ 1,  .8,  .8}0.030  \\
    NvDiffRec~\cite{NvDiffRec} & 0.045  & 0.101  & 0.048  & 0.060  & 0.061  & 0.100  & 0.040  & 0.191  & 0.081  \\
    MobileNeRF~\cite{mobilenerf} & 0.025  & 0.077  & 0.048  & 0.050  & 0.025  & 0.092  & 0.032  & 0.145  & 0.062  \\
    NeRF2Mesh~\cite{nerf2mesh} & 0.046  & 0.084  & 0.045  & 0.060  & 0.047  & 0.107  & 0.042  & 0.145  & 0.072  \\
    Ours  & \cellcolor[rgb]{ 1,  .949,  .8}0.018  & \cellcolor[rgb]{ .973,  .796,  .678}0.045  & \cellcolor[rgb]{ 1,  .949,  .8}0.027  &\cellcolor[rgb]{ 1,  .949,  .8}0.024  &\cellcolor[rgb]{ 1,  .949,  .8}0.024  & \cellcolor[rgb]{ 1,  .949,  .8}0.041  & \cellcolor[rgb]{ 1,  .949,  .8}0.013  & \cellcolor[rgb]{ .973,  .796,  .678}0.125  & \cellcolor[rgb]{ 1,  .949,  .8}0.040  \\
    \hline
    Ours (Mobile) & \cellcolor[rgb]{ 1,  .949,  .8}0.018  & \cellcolor[rgb]{ 1,  .949,  .8}0.047  & 0.028  & 0.027  & 0.027  & 0.044  & 0.015  & \cellcolor[rgb]{ 1,  .949,  .8}0.130  & 0.042  \\
    \bottomrule
    \end{tabular}%
    \vspace{3pt}
    \caption{\textbf{LPIPS$\downarrow$ on NeRF Synthetic Dataset.} Red represents the optimal, orange represents the second best, and yellow represents the third.}
  \label{tab:lpipsnerf}%
\end{table}%

\begin{table}[!tbp]
  \centering
    \small
    \setlength{\tabcolsep}{1.5mm}
    \begin{tabular}{c|ccccccc}
    \toprule
          & Ball & Teapot & Helmet & Toaster & Coffee & Car   & Mean \\
    \hline
    Ref-NeRF~\cite{refnerf} & \cellcolor[rgb]{ 1,  .8,  .8}47.46 & \cellcolor[rgb]{ 1,  .8,  .8}47.90 & \cellcolor[rgb]{ 1,  .949,  .8}29.68  & \cellcolor[rgb]{ 1,  .8,  .8}25.70  & \cellcolor[rgb]{ .973,  .796,  .678}34.21  & \cellcolor[rgb]{ 1,  .949,  .8}30.82  & \cellcolor[rgb]{ .973,  .796,  .678}35.96  \\
    \hline
    3DGS~\cite{3DGS} & 27.69  & 45.68  & 28.32  & 20.99  & 32.32 & 27.24 & 30.37  \\
    NvDiffRec~\cite{NvDiffRec} & 23.85 & 40.38 & 27.58  & 24.12  & 30.52  & 27.85  & 29.05  \\
    MobileNeRF~\cite{mobilenerf} & 17.43 & 42.91 & 25.91  & 19.82  & 29.29  & 24.38  & 26.62  \\
    NeRF2Mesh~\cite{nerf2mesh}& 20.57 & 40.10 & 25.02  & 18.93  & 31.29  & 25.86  & 26.96  \\
    Ours  & \cellcolor[rgb]{ .973,  .796,  .678}39.63 & \cellcolor[rgb]{ .973,  .796,  .678}47.06 & \cellcolor[rgb]{ 1,  .8,  .8}38.78 & \cellcolor[rgb]{ .973,  .796,  .678}24.95 & \cellcolor[rgb]{ 1,  .8,  .8}34.25 & \cellcolor[rgb]{ 1,  .8,  .8}31.45 & \cellcolor[rgb]{ 1,  .8,  .8}36.02  \\
    \hline
    Ours (Mobile) & \cellcolor[rgb]{ 1,  .949,  .8}39.38 & \cellcolor[rgb]{ 1,  .949,  .8}46.71 & \cellcolor[rgb]{ .973,  .796,  .678}38.52  & \cellcolor[rgb]{ 1,  .949,  .8}24.90  & \cellcolor[rgb]{ 1,  .949,  .8}34.12  & \cellcolor[rgb]{ .973,  .796,  .678}31.33  & \cellcolor[rgb]{ 1,  .949,  .8}35.83  \\
    \bottomrule
    \end{tabular}%
    \vspace{3pt}
  \caption{\textbf{PSNR$\uparrow$ on Shiny Blender Dataset.} Red represents the optimal, orange represents the second best, and yellow represents the third.}
  \label{tab:psnrshiny}%
\end{table}%

\begin{table}[!tbp]
  \centering
    \small
        \setlength{\tabcolsep}{1.5mm}
    \begin{tabular}{c|ccccccc}
    \toprule
          & Ball & Teapot & Helmet & Toaster & Coffee & Car   & Mean \\
    \hline
    Ref-NeRF~\cite{refnerf} & \cellcolor[rgb]{ 1,  .8,  .8}0.995 & \cellcolor[rgb]{ 1,  .8,  .8}0.998 & \cellcolor[rgb]{ 1,  .949,  .8}0.958  & \cellcolor[rgb]{ 1,  .949,  .8}0.922  & \cellcolor[rgb]{ 1,  .949,  .8}0.974  & \cellcolor[rgb]{ 1,  .949,  .8}0.955  &\cellcolor[rgb]{ 1,  .949,  .8}0.967  \\
    \hline
    3DGS~\cite{3DGS} & 0.937 & \cellcolor[rgb]{ .973,  .796,  .678}0.996 & 0.951  & 0.895 & 0.971  & 0.930  & 0.947  \\
    NvDiffRec~\cite{NvDiffRec} & 0.888 & \cellcolor[rgb]{ 1,  .949,  .8}0.993 & 0.939  & 0.903 & 0.958  & 0.947  & 0.938  \\
    MobileNeRF~\cite{mobilenerf} & 0.775 & \cellcolor[rgb]{ .973,  .796,  .678}0.996 & 0.892  & 0.784  & 0.954  & 0.897  & 0.883  \\
    NeRF2Mesh~\cite{nerf2mesh}& 0.805 & 0.987 & 0.912  & 0.806  & 0.958  & 0.909  & 0.896  \\
    Ours  & \cellcolor[rgb]{ .973,  .796,  .678}0.994 & \cellcolor[rgb]{ 1,  .8,  .8}0.998 & \cellcolor[rgb]{ 1,  .8,  .8}0.993 & \cellcolor[rgb]{ 1,  .8,  .8}0.957 & \cellcolor[rgb]{ 1,  .8,  .8}0.977 & \cellcolor[rgb]{ 1,  .8,  .8}0.977 & \cellcolor[rgb]{ 1,  .8,  .8}0.983  \\
    \hline
    Ours (Mobile) & \cellcolor[rgb]{ 1,  .949,  .8}0.992 & \cellcolor[rgb]{ 1,  .8,  .8}0.998 & \cellcolor[rgb]{ .973,  .796,  .678}0.991 & \cellcolor[rgb]{ .973,  .796,  .678}0.954 & \cellcolor[rgb]{ .973,  .796,  .678}0.976 & \cellcolor[rgb]{ .973,  .796,  .678}0.976 & \cellcolor[rgb]{ .973,  .796,  .678}  0.981  \\
    \bottomrule
    \end{tabular}%
    \vspace{3pt}
      \caption{\textbf{SSIM$\uparrow$ on Shiny Blender Dataset.} Red represents the optimal, orange represents the second best, and yellow represents the third.}

  \label{tab:ssimshiny}%
\end{table}%

\begin{table}[!tbp]
  \centering
    \small
        \setlength{\tabcolsep}{1.5mm}
    \begin{tabular}{c|ccccccc}
    \toprule
          & Ball & Teapot & Helmet & Toaster & Coffee & Car   & Mean \\
    \hline
    Ref-NeRF~\cite{refnerf}& \cellcolor[rgb]{ 1,  .8,  .8}0.059 & \cellcolor[rgb]{ 1,  .8,  .8}0.004 & \cellcolor[rgb]{ 1,  .949,  .8}0.075  & \cellcolor[rgb]{ 1,  .949,  .8}0.095  & \cellcolor[rgb]{ .973,  .796,  .678}0.078  & \cellcolor[rgb]{ 1,  .949,  .8}0.041  & \cellcolor[rgb]{ 1,  .949,  .8}0.058  \\
    \hline
    3DGS~\cite{3DGS} & 0.161  & \cellcolor[rgb]{ .973,  .796,  .678}0.007  & 0.079  & 0.126  & \cellcolor[rgb]{ .973,  .796,  .678}0.078 & 0.047 & 0.083 \\
    NvDiffRec~\cite{NvDiffRec}& 0.200 & 0.021 & 0.123  & 0.152  & \cellcolor[rgb]{ 1,  .949,  .8}0.111  & 0.060  & 0.111  \\
    MobileNeRF~\cite{mobilenerf} & 0.330 & \cellcolor[rgb]{ 1,  .949,  .8}0.013 & 0.175  & 0.244  & 0.112  & 0.105  & 0.163  \\
    NeRF2Mesh~\cite{nerf2mesh} & 0.321 & 0.041 & 0.193  & 0.250  & 0.122  & 0.090  & 0.170  \\
    Ours  &\cellcolor[rgb]{ 1,  .949,  .8}0.064 & \cellcolor[rgb]{ 1,  .8,  .8}0.004 & \cellcolor[rgb]{ 1,  .8,  .8}0.015 & \cellcolor[rgb]{ 1,  .8,  .8}0.054 & \cellcolor[rgb]{ 1,  .8,  .8}0.077 & \cellcolor[rgb]{ 1,  .8,  .8}0.024 & \cellcolor[rgb]{ 1,  .8,  .8}0.040 \\
    \hline
    Ours (Mobile) &\cellcolor[rgb]{ .973,  .796,  .678}0.060 & \cellcolor[rgb]{ 1,  .8,  .8}0.004 & \cellcolor[rgb]{ .973,  .796,  .678}0.021 & \cellcolor[rgb]{ .973,  .796,  .678}0.061 & \cellcolor[rgb]{ 1,  .8,  .8}0.077 & \cellcolor[rgb]{ .973,  .796,  .678}0.025 & \cellcolor[rgb]{ .973,  .796,  .678}0.041   \\
    \bottomrule
    \end{tabular}%
    \vspace{3pt}
    \caption{\textbf{LPIPS$\downarrow$ on Shiny Blender Dataset.} Red represents the optimal, orange represents the second best, and yellow represents the third.}
  \label{tab:lpipsshiny}%
\end{table}%

\begin{table}[!tbp]
  \centering
    \small
    \setlength{\tabcolsep}{3mm}

    \begin{tabular}{c|cccr}
    \toprule
          & Sedan & Toycar & Gardenspheres & \multicolumn{1}{c}{Mean} \\
    \hline
    Ref-NeRF~\cite{refnerf} & \cellcolor[rgb]{ 1,  .8,  .8}25.65  & \cellcolor[rgb]{ .973,  .796,  .678}24.25  & \cellcolor[rgb]{ 1,  .8,  .8}23.46  & \cellcolor[rgb]{ 1,  .8,  .8}24.45  \\
    \hline
    3DGS~\cite{3DGS} & \cellcolor[rgb]{ .973,  .796,  .678}25.48  & \cellcolor[rgb]{ 1,  .8,  .8}24.35  & \cellcolor[rgb]{ .973,  .796,  .678}22.34  & \cellcolor[rgb]{ .973,  .796,  .678}24.06  \\
    MobileNeRF~\cite{mobilenerf} & \cellcolor[rgb]{ 1,  .949,  .8}23.54  & \cellcolor[rgb]{ 1,  .949,  .8}23.71  & 21.62  & \cellcolor[rgb]{ 1,  .949,  .8}22.96  \\
    NeRF2Mesh~\cite{nerf2mesh} & 22.83  & 23.70  & 21.59  & 22.71  \\
    Ours  & 22.83  & 23.00  & \cellcolor[rgb]{ 1,  .949,  .8}21.66  & 22.50  \\
    \hline
    Ours (Mobile) & 21.88  & 22.85  & 21.61  & 22.11  \\
    \bottomrule
    \end{tabular}%
    \vspace{3pt}
    \caption{\textbf{PSNR$\uparrow$ on Real Captured Dataset.} Red represents the optimal, orange represents the second best, and yellow represents the third.}
  \label{tab:psnrreal}%
\end{table}%

\begin{table}[!tbp]
  \centering
    \small
    \setlength{\tabcolsep}{3mm}

    \begin{tabular}{c|cccc}
    \toprule
          & Sedan & Toycar & Gardenspheres & Mean \\
    \hline
    Ref-NeRF~\cite{refnerf} & \cellcolor[rgb]{ .973,  .796,  .678}0.720  & \cellcolor[rgb]{ 1,  .8,  .8}0.674  & \cellcolor[rgb]{ 1,  .8,  .8}0.601  & \cellcolor[rgb]{ 1,  .8,  .8}0.665  \\
    \hline
    3DGS~\cite{3DGS} & \cellcolor[rgb]{ 1,  .8,  .8}0.726  & \cellcolor[rgb]{ .973,  .796,  .678}0.662  & \cellcolor[rgb]{ .973,  .796,  .678}0.596  & \cellcolor[rgb]{ .973,  .796,  .678}0.661  \\
    MobileNeRF~\cite{mobilenerf} & 0.531  & 0.545  & 0.404  & 0.493  \\
    NeRF2Mesh~\cite{nerf2mesh} & 0.556  & 0.588  & 0.426  & 0.523  \\
    Ours  &  \cellcolor[rgb]{ 1,  .949,  .8}0.584  & \cellcolor[rgb]{ 1,  .949,  .8} 0.624  &  \cellcolor[rgb]{ 1,  .949,  .8}0.586  &  \cellcolor[rgb]{ 1,  .949,  .8}0.598  \\
    \hline
    Ours (Mobile) & 0.532  & 0.547  & 0.495  & 0.524  \\
    \bottomrule
    \end{tabular}%
    \vspace{3pt}
      \caption{\textbf{SSIM$\uparrow$ on Real Captured Dataset.} Red represents the optimal, orange represents the second best, and yellow represents the third.}

  \label{tab:ssimreal}%
\end{table}%

\begin{table}[!tbp]
  \centering
    \small
    \setlength{\tabcolsep}{3mm}

    \begin{tabular}{c|cccc}
    \toprule
          & Sedan & Toycar & Gardenspheres & Mean \\
    \hline
    Ref-NeRF~\cite{refnerf} & \cellcolor[rgb]{ 1,  .8,  .8}0.119  & \cellcolor[rgb]{ 1,  .8,  .8}0.168  & \cellcolor[rgb]{ 1,  .8,  .8}0.138  & \cellcolor[rgb]{ 1,  .8,  .8}0.142  \\
    \hline
    3DGS~\cite{3DGS} & \cellcolor[rgb]{ .973,  .796,  .678}0.301  & \cellcolor[rgb]{ .973,  .796,  .678}0.234  & \cellcolor[rgb]{ .973,  .796,  .678}0.243  & \cellcolor[rgb]{ .973,  .796,  .678}0.259  \\
    MobileNeRF~\cite{mobilenerf} & 0.494  & 0.392  & 0.404  & 0.430  \\
    NeRF2Mesh~\cite{nerf2mesh} & 0.487  & 0.343  & 0.427  & 0.419  \\
    Ours  & \cellcolor[rgb]{ 1,  .949,  .8}0.431  & \cellcolor[rgb]{ 1,  .949,  .8}0.271  & \cellcolor[rgb]{ 1,  .949,  .8}0.273  & \cellcolor[rgb]{ 1,  .949,  .8}0.325  \\
    \hline
    Ours (Mobile) & 0.531  & 0.370  & 0.369  & 0.423  \\
    
    \bottomrule
    \end{tabular}%
    \vspace{3pt}
      \caption{\textbf{LPIPS$\downarrow$ on Real Captured Dataset.} Red represents the optimal, orange represents the second best, and yellow represents the third.}
  \label{tab:lpipsreal}%
\end{table}%

\begin{table}[h]
  \centering
    \small
    \begin{tabular}{c|ccc}
    \toprule
          & NeRF Synthetic & Shiny Blender & Real Captured \\
    \hline
    MacBook Pro & 120.00   & 120.00   & 120.00 \\
    iPhone12 & 60.00    & 60.00    & 60.00 \\
    iPad Air3 & 60.00    & 60.00    & 60.00 \\
    Legion Y-7000 & 48.50  & 48.00    & 32.12 \\
    NVIDIA GeForce RTX 3090 & 219.44 & 223.55 & 145.54 \\
    \bottomrule
    \end{tabular}%
    \vspace{3pt}
    \caption{\textbf{Detailed Rendering Efficiency.} We also test the FPS (frames per second) of Ours (Mobile) on the Shiny Blender dataset and Real Captured dataset. On MacBook Pro, iPhone 12, and iPad Air3, our method achieves the maximum screen refresh rate among all datasets.}
  \label{tab:oursfps}%
\end{table}%

\subsection{Rendering Efficiency}
\label{sec:renderingefficiency}

\boldparagraph{Computational efficiency} During the mobile rendering stage, the main bottleneck in rendering speed lies in the process of querying the $f_\psi$ to obtain $\bc_s$. Since our $f_\psi$ only has 2 layers with 64 widths, we achieve rendering in real-time on mobile devices. We compare the rendering speed with the baseline methods~\cite{mobilenerf,nerf2mesh} on the same set of hardware devices in~\cref{tab:fps} of our main text. For the baseline methods~\cite{mobilenerf,nerf2mesh}, we use their official web demo\footnote{https://mobile-nerf.github.io/,https://me.kiui.moe/nerf2mesh/} for comparison. Additionally, since they do not provide demos for the Shiny Blender dataset and the Real Captured dataset, we only compare the NeRF Synthetic dataset. In addition, we also test the rendering speed of our method on the Shiny Blender dataset and Real Captured dataset on various devices, as shown in~\cref{tab:oursfps}.

\subsection{Ablation Study}
\label{sec:abl}
\begin{table}[t]
\small
      \setlength{\tabcolsep}{2mm}
  \centering
    \begin{tabular}{cc|ccc}
    \toprule
          &       & PSNR$\uparrow$  & SSIM$\uparrow$  & LPIPS$\downarrow$ \\
    \hline
    \multirow{2}[1]{*}{Toaster} & w/o SSIM & 23.92 & 0.866 & 0.184 \\
          & Ours & 24.95 & 0.957 & 0.054 \\
    \multirow{2}[1]{*}{Materials} & w/o SSIM & 28.59 & 0.928 & 0.079 \\
          & Ours & 29.58 & 0.962 & 0.041 \\
    \bottomrule
    \end{tabular}%
    \vspace{3pt}
  \caption{\textbf{Ablation of Loss Function.}}
  \label{tab:ablssimtab}%

\end{table}%

\boldparagraph{Loss Function} We found that introducing the SSIM~\cite{ssim} loss greatly improves the subjective quality. Moreover, the PSNR metric also improves with the help of SSIM, as shown in~\cref{tab:ablssimtab}.

\begin{table}[h]
  \centering
    \setlength{\tabcolsep}{7.5mm}

    \begin{tabular}{c|cc}
    \toprule
          & w/o $\Delta$    & Ours \\
    \hline
    Car   & 26.84  & 31.45  \\
    Helmet & 33.93  & 38.78  \\
    Toaster & 23.01  & 24.95  \\
    Coffee & 27.80  & 34.25  \\
    \bottomrule
    \end{tabular}%
    \vspace{3pt}
      \caption{\textbf{Ablation of Mesh Optimization.} "w/o $\Delta$" refers to the version without mesh optimization. Reported in PSNR.}
  \label{tab:ablopti}%
\end{table}%

\boldparagraph{Mesh Optimization} Since the initial mesh provided by Ref-Neus~\cite{refneus} is already of sufficient quality, we also attempt to remove the optimization of the mesh ($\Delta \bn_\bv$ and $\Delta \bv$) on the Shiny Blender dataset. However, we found that this results in a decrease in rendering quality, as shown in~\cref{tab:ablopti}.

\subsection{Scene Editing}
\label{sec:sceneediting}
\begin{figure}[h] %
    \begin{subfigure}[t]{0.25\linewidth} %
        \centering
        \includegraphics[width=1\linewidth]{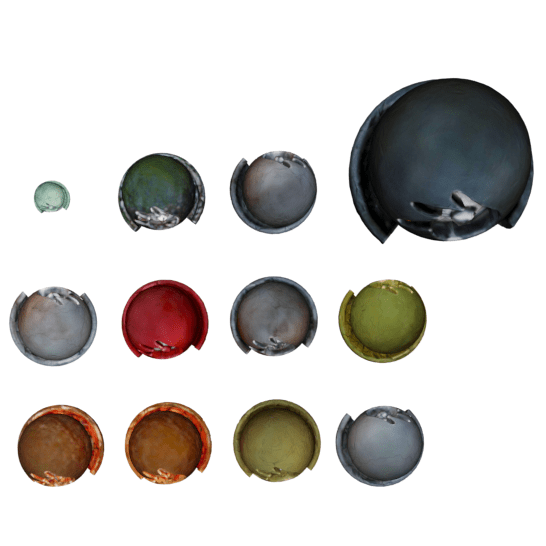} 
        \caption{Geometry Editing.} %
    \end{subfigure}%
    \begin{subfigure}[t]{0.25\linewidth} %
        \centering
        \includegraphics[width=1\linewidth]{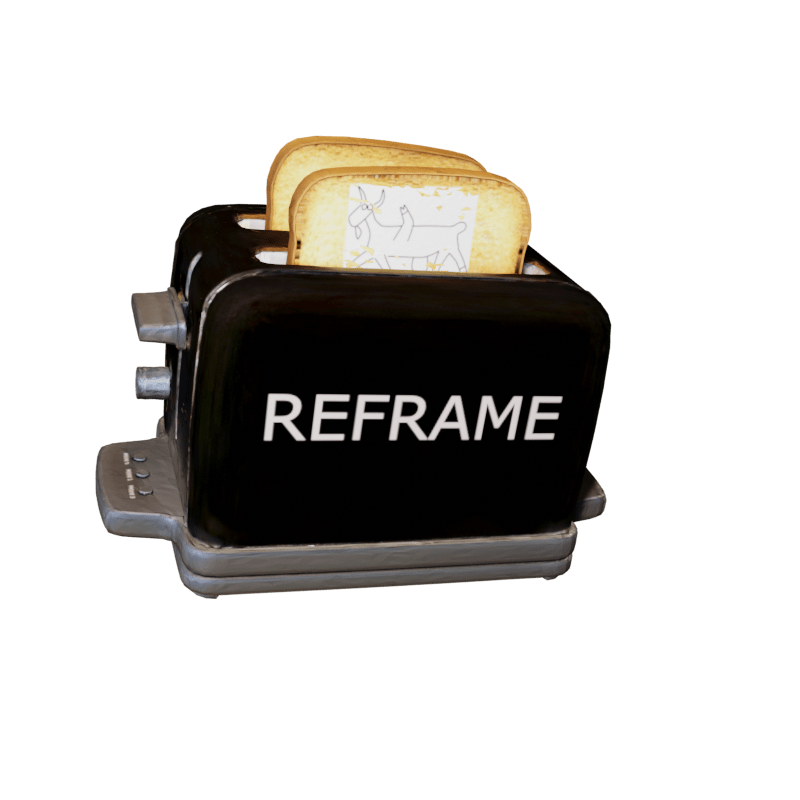} 
        \caption{Appearance Editing.} %
    \end{subfigure}%
    \begin{subfigure}[t]{0.25\linewidth} %
        \centering
        \includegraphics[width=1\linewidth]{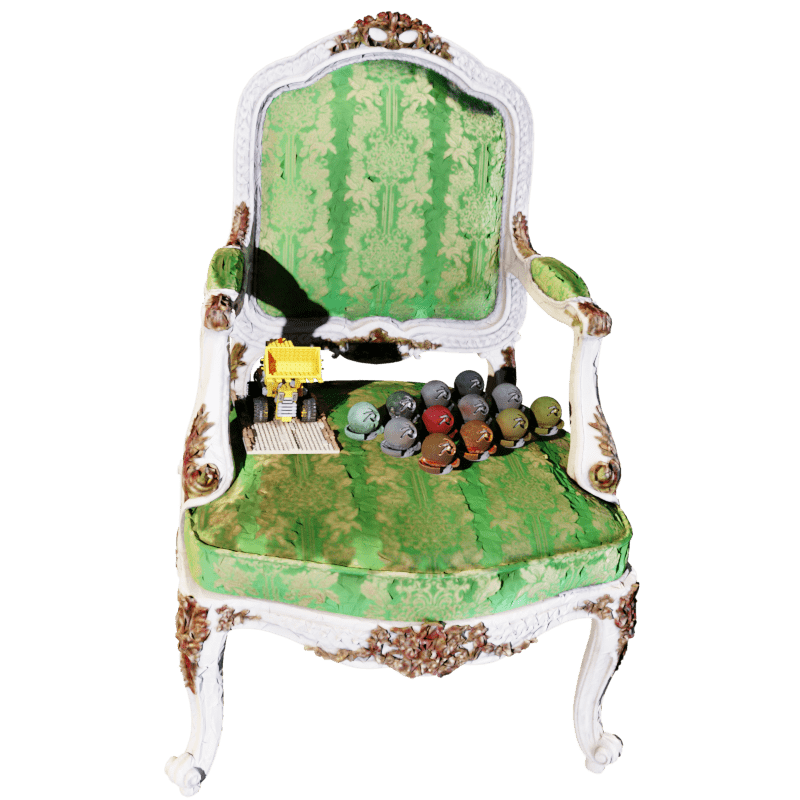} 
        \caption{Composition.} %
    \end{subfigure}%
    \begin{subfigure}[t]{0.25\linewidth} %
        \centering
        \includegraphics[width=1\linewidth]{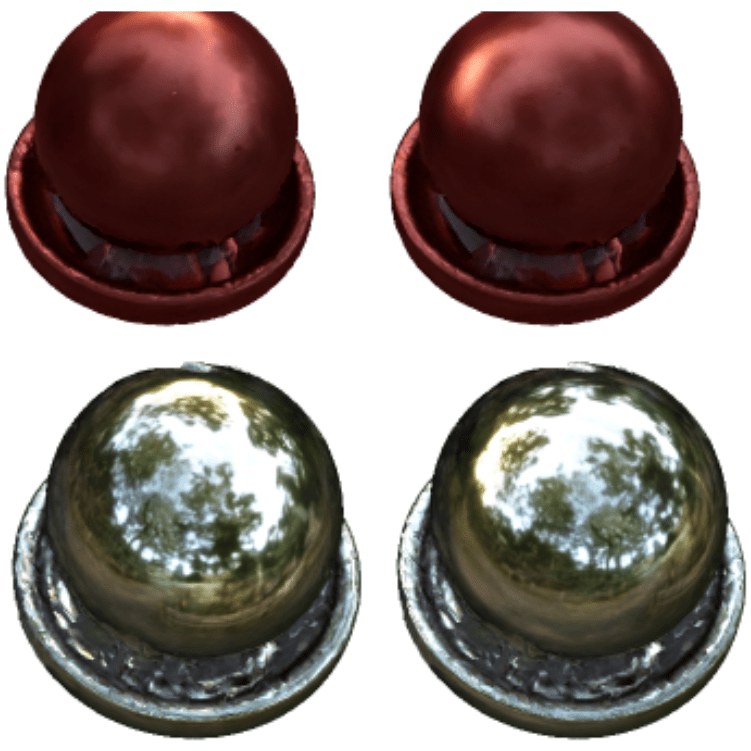} 
        \caption{Relighting.} %
    \end{subfigure}%
    \caption{\textbf{Scene Editing.} We can perform some scene editing tasks.}
  \label{fig:sceneedit}
\end{figure}

Having decoupled the environmental information, geometry, and appearance of the scene, we can perform some scene editing tasks, as demonstrated in~\cref{fig:sceneedit}.

\end{document}